\documentclass[10pt,twocolumn,letterpaper]{article}

\pdfoutput=1

\usepackage{arxiv}

\usepackage{times}
\usepackage{epsfig}
\usepackage{graphicx}
\usepackage{amsmath}
\usepackage{amssymb}
\usepackage{enumitem}
\usepackage{makecell}
\usepackage{multirow}
\usepackage{multicol}
\usepackage{rotating}
\usepackage{algorithm}
\usepackage{algpseudocode} 
\usepackage{graphicx}
\usepackage[outdir=./]{epstopdf}
\usepackage[dvipsnames]{xcolor}
\usepackage{comment}
\usepackage{footmisc}
\usepackage{booktabs}
\usepackage[toc,page]{appendix}

\usepackage[breaklinks=true,bookmarks=false]{hyperref}
\usepackage{url}

\usepackage{pifont}
\newcommand{\xmark}{\ding{55}}

\newcommand{\figref}[1]{Fig.~\ref{#1}}
\newcommand{\tabref}[1]{Tab.~\ref{#1}}

\newcommand{\AlgRef}[1]{Algorithm~\ref{#1}}

\newcommand{\tabincell}[2]{\begin{tabular}{@{}#1@{}}#2\end{tabular}}
\newcommand{\argmin}{\mathop{\mathrm{arg\,min}}}

\def\iccvPaperID{8035} 

\begin{document}

\title{When to Prune? A Policy towards Early Structural Pruning}
\author{Maying Shen, Pavlo Molchanov, Hongxu Yin, Jose M. Alvarez\\
NVIDIA\\
{\tt\small \{mshen, pmolchanov, dannyy, josea\}@nvidia.com}}


\maketitle

\begin{abstract}
Pruning enables appealing reductions in network memory footprint and time complexity. Conventional post-training pruning techniques lean towards efficient inference while overlooking the heavy computation for training. Recent exploration of pre-training pruning at initialization hints on training cost reduction via pruning, but suffers noticeable performance degradation. We attempt to combine the benefits of both directions and propose a policy that prunes as early as possible during training without hurting performance. Instead of pruning at initialization, our method exploits initial dense training for few epochs to quickly guide the architecture, while constantly evaluating dominant sub-networks via neuron importance ranking. This unveils dominant sub-networks whose structures turn stable, allowing conventional pruning to be pushed earlier into the training. To do this early, we further introduce an Early Pruning Indicator (EPI) that relies on sub-network architectural similarity and quickly triggers pruning when the sub-network's architecture stabilizes. Through extensive experiments on ImageNet, we show that EPI empowers a quick tracking of early training epochs suitable for pruning, offering same efficacy as an otherwise ``oracle'' grid-search that scans through epochs and requires orders of magnitude more compute. Our method yields $1.4\%$ top-1 accuracy boost over state-of-the-art pruning counterparts, cuts down training cost on GPU by $2.4\times$, hence offers a new efficiency-accuracy boundary for network pruning during training.


\end{abstract}

\section{Introduction}
\label{intro}
The Success of convolutional neural networks (CNNs) fuels the recent progress in computer vision, boosting up performance for classification, detection, and segmentation tasks~\cite{he2015deep,Simonyan15,szegedy2015going}. While enjoying the accuracy benefits CNNs bring, a simultaneous increase in network complexity imposes higher memory footprint and computing power consumption, making deployment of CNNs on resource-constrained devices a challenging task~\cite{chamnet,molchanov2019importance,molchanov2016pruning}. In lieu of computation-intensive networks, recent work turn to compression techniques for efficient models leveraging pruning~\cite{alvarez2016learning, han2015deep,liu2018rethinking,molchanov2019importance}, quantization~\cite{cai2020zeroq,wang2020apq,zhu2016trained}, knowledge distillation~\cite{hinton2015distilling,mullapudi2019online,yin2020dreaming}, neural architecture search~\cite{tan2019mnasnet,vahdat2020unas,wu2019fbnet}, and architecture redesigns~\cite{howard2017mobilenets,ma2018shufflenet,tan2019efficientnet}. Among these, pruning demonstrates to be a widely adopted method that compresses pre-trained models before deployment. The primary goal of pruning aims to remove insignificant network parameters without impacting accuracy. In particular, structural pruning removes entire filters (or neurons) as such the resulting structural sparsity benefits legacy off-the-shelf platforms, \eg, CPUs, DSPs, and GPUs. 

\begin{figure}[!t]
\vspace{-2mm}
    \centering
    \includegraphics[ scale=0.45]{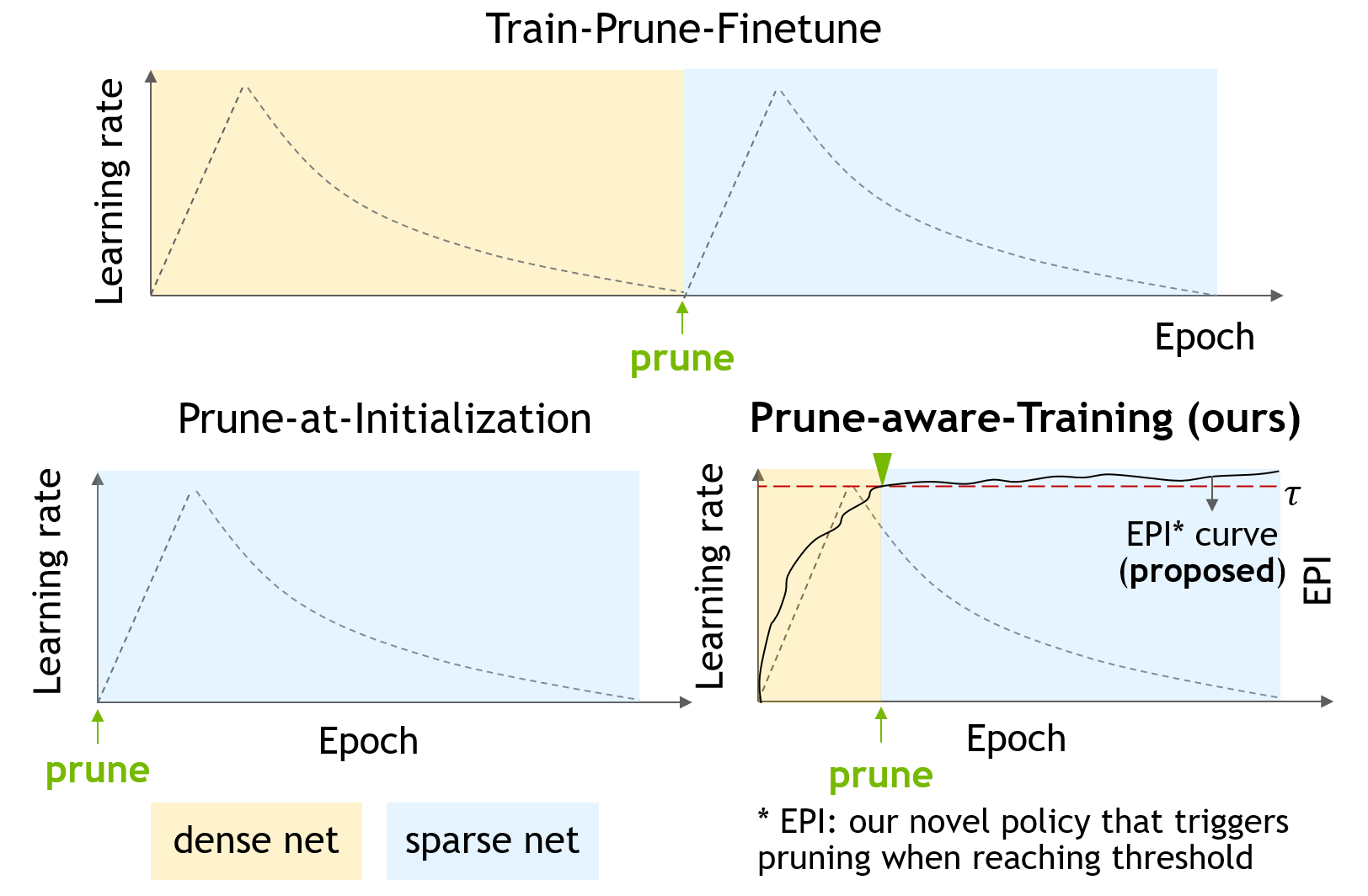}
    \caption{Pruning paradigm overview. \textbf{\emph{Train-Prune-Finetune}} prunes after training, effective but costs additional training time; 
    \textbf{\emph{Prune-at-Initialization}} prunes right before training towards a smaller network, cuts down on training time 
    but suffers notable performance degradation;
    \textbf{\emph{Pruning-aware-Training} (ours)} prunes during training aiming at benefits from two worlds. It governs on post-pruning performance while aiming to minimize training time, via a new policy around Early Pruning Indicator (EPI) that signals an early optimal point to start pruning during training.
    }
    \vspace{-5mm}
    \label{fig:diff_with_different_prune_point}
\end{figure}

In general, network pruning involves three key steps: (i) original training of a dense model for high accuracy, (ii) pruning away insignificant weights to remove redundancy, and finally (iii) fine-tuning the pruned model to recover performance~\cite{han2015deep, luo2017thinet, molchanov2016pruning}. Despite remarkable compactness delivered by the last two steps, the original training of an over-parameterized network remains mostly untouched. Such approaches require a twice long time (resource) as an original training recipe given similar required computes for fine-tuning, making the entire pipeline slow, sometimes infeasible. For example, a very recent breakthrough in language modeling, a GPT-$3$ model~\cite{brown2020language}, requires millions of dollars (more than $300$ NVIDIA V100 GPU years) just for the initial training. Already aware of post-training redundancy, an interesting question arises - can we somehow prune a network during its initial training, as such the resulting sparsity can (i) immediately benefit training and (ii) save us from the costly additional fine-tuning upon training ends? 

One intuitive and ideal solution to this problem is pruning the network right at initialization even before training starts. The intriguing observation from the Lottery Ticket Hypothesis~\cite{frankle2019lottery} hints potential to this task: it shows the (i) existence of small sub-models, identifiable via pruning, within a large dense model, that can (ii) be trained in isolation to achieve the same accuracy as its dense counterpart~\cite{frankle2019lottery,liu2018rethinking}.
This field has quickly evolved and recent approaches have enhanced policy for optimal sub-network at the initialization by preserving the loss or the gradient flow~\cite{de2020force,lee2018snip,Wang2020grasp}. Despite rapid progress, the approaches of sub-network identification at the initialization remain challenging and still suffer noticeable accuracy loss~\cite{frankle2019linear,gale2019state}. 

Instead of \textit{zero} training, in this work, we showcase the benefits and practicability of pruning \textit{early} during training. Doing so allows one to (i) save compute by training only pruned models most of the time, (ii) alleviate any extra fine-tuning by aligning the process with original training, while (iii) suppressing accuracy loss by moving slightly later into the training regime for pruning guidance. We name this approach~\textbf{pruning-aware-training} (PaT). As shown in Fig.~\ref{fig:diff_with_different_prune_point}, unlike pruning-at-initialization, PaT takes full advantage of early-stage dense model training that is beneficial for rapid learning and optimal architecture exploration~\cite{achille2018critical,Frankle2020earlyphase}, while aiming to identify the best sub-network as early as possible, rather than waiting till training ends as in conventional pruning. 

The key of benefiting from the training efficiency of PaT and saving training time relies on finding an early yet eligible point during training to start pruning. Existing methods that perform pruning during training~\cite{alvarez2016learning,frankle2019linear, lym2019prunetrain,oyedotun2020structured} have shown the efficacy of this direction by reducing the turn-around time. However, in most cases a fixed initial interval for pruning is set heuristically, or post-training statistics are required.
In this work we focus on understanding how the starting point of pruning can be set automatically.

We start by analyzing in depth the evolution of pruned architectures via performing trimming across all epochs rigorously and compare their suitability for pruning. Though laborious, this oracle estimate offers key insights on pruning during training. We observe an important property: agnostic of magnitude or gradient criterion, (i) pruning at early epochs results in different final architectures, but (ii) dominant architecture emerges within just a few epochs and stabilizes thereafter till training ends, allowing conventional pruning to be pushed earlier into the training.

Amid such property we further propose a novel metric, called Early Pruning Indicator (EPI), that estimates the structure similarity between networks resulted in pruning at consecutive epochs of the same base model. Given intrinsic access to model weights and gradients during training, EPI can be calculated very efficiently alongside initial training without bells and whistles, while helping avoid the otherwise lengthy grid search for starting epochs. Once the resulting pruning structure will not vary between epochs we argue and demonstrate it is safe to prune.
As prior work~\cite{liu2018rethinking} and we observe, structural pruning acts as an architecture search and tries to find the optimal number of neurons per layer. Therefore, we hypothesize that pruning can be performed as soon as the architecture of the dominant sub-network becomes stable. 
We demonstrate that the proposed metric works across varying network architectures, pruning ratios, delivering consistent reductions in training time.

Our main contributions are as follows: 

\begin{itemize}[topsep=1pt,itemsep=1pt,partopsep=0pt, parsep=0pt,leftmargin=\labelwidth]
\item We propose a novel metric called Early Pruning Indicator (EPI) that indicates an early point to start pruning during training. Our metric enables training to benefit from sparsity, significantly reducing training resources with minor accuracy drop.
\item We demonstrate that for structural pruning (output channel pruning), initial dense training fuels accuracy boosts. Augmented by EPI, our pruning-aware-training outperforms pruning-at-initialization alternatives by a large margin. 
\item We show that EPI is agnostic to the pruning method used by showing efficacy for both magnitude-based and gradient-based pruning, enabling a new state-of-the-art boundary for training speedup through in-situ pruning.
\end{itemize}

\section{Related Work}
\label{Sect:related_work}

\noindent\textbf{Network pruning.} Mainstream pruning methods can be divided into three categories depending on when pruning is performed: 1) train-prune-finetune, 2) prune at initialization, and 3) prune while training. 

The first group, train-prune-finetune, performs pruning on a densely pre-trained network and then, fine-tune the resulting structure to recover the performance loss caused by pruning. There are many methods aiming at preserving the final accuracy~\cite{he2020learning, molchanov2019importance, molchanov2016pruning} and minimize the output change of each layer~\cite{he2019filter, luo2017thinet}. A key focus of these work resides in identifying redundant connections whose removal brings the least perturbation to the overall performance. While enabling plausible performance and improving efficiency at test time, the aforementioned approaches cannot yet bring any efficiency benefit to training. Quite in contrary, most recipes result in nearly doubled training time amid the requirement for lengthy fine-tuning. 

Prune at initialization methods, backed by the Lottery Ticket Hypothesis~\cite{frankle2019lottery}, question the necessity of dense training for performance convergence~\cite{liu2018rethinking}. A forerunner in this group is SNIP~\cite{lee2018snip}, an approach that identifies a trainable sub-network at initialization. Subsequently, other methods such as FORCE~\cite{de2020force}, GraSP~\cite{Wang2020grasp}, or SynFlow~\cite{tanaka2020pruning} have been proposed to improve performance. These methods make the training more efficient as they only train the sparse network. However, the reliability of pruning at initialization remains unsatisfactory facing inevitable performance gaps~\cite{frankle2019linear}.

Prune while training methods rest in the middle by finding a trade-off between training efficiency and final accuracy. Literature falls under two streams towards this task: a) regularization-based methods that encourage sparsity during training~\cite{alvarez2016learning,gao2019vacl,lym2019prunetrain}, and b) sub-ticket selection methods via saliency that discard redundancy~\cite{GPWP,goh1994pruning,he2020learning}. Our work belongs to the latter given its efficacy to quickly enforce a pruning ratio and ease-of-control during training. Under this realm, one line of work learns sub-networks during training~\cite{alvarez2016learning, lym2019prunetrain, oyedotun2020structured}. Others, such as Frankle \etal \cite{frankle2019linear}, study the need of few training iterations before pruning in order to maximize the performance. These methods struggle to automatically identify the starting points at which pruning can be performed, while heavily relying on hand-crafted or post-training heuristics for decision making. 

\noindent \textbf{Network similarity.} Our policy explores the network similarity of two sub-networks resulted from pruning. A comprehensive review of network similarity measures was presented in \cite{netsimilarity}. These methods aim at comparing the representation between two fully trained models with different initialization, hence are not applicable to in-training gauging for pruning where weights and dominant architectures are both changing. To get the structural similarity for pruning, we focus on the number of remaining neurons across layers in a network when comparing with another one. The difference between these skeletons using coefficients can be directly measured by Spearman's \cite{spearman1961proof} and Kendall's tau \cite{kendall1938new} rank correlation. However, these rank correlation metrics take into account the specific ranking and, more importantly, they would rely on all neurons in the network. Thus, they provide the same value for all pruning ratios.

Of particular relevance to this work is~\cite{frankle2019linear} by Frankle~\etal that proposes an approach to measure the instability of a network structure to understand pruning viability. One noticeable finding by this work shows that the best time to perform iterative magnitude pruning tends to be after some initial training. Interestingly it identifies a relationship between the model instability and the accuracy of the pruned network, though the instability measure proposed by the method can only be measured \textit{after training is completed}, hence remains insufficient to directly signal when to start pruning during training. 

\section{Method}

We next elaborate our early pruning algorithm in details.

\subsection{Objective Function}
\begin{algorithm}[!t]
    \caption{Iterative pruning within one epoch}
    \label{alg:iterative-pruning-algo}
    \begin{algorithmic}[1]
        \State For prune ratio $\alpha$, schedule the number of neurons to prune per step for $S$ steps via exponential scheduler~\cite{de2020force}, forming $\textbf{m}\in \mathbb{R}^{S}$ 
        \While{ $| \mathcal{P} | \leq \alpha | \mathcal{F} |$}
            \State Average importance calculated by \eqref{mag-based-importance} or \eqref{grad-based-importance} over multiple minibatches
            \State $\mathcal{P}$ is the indices of $\textbf{m}_i$ bottom-ranked neurons
            \State $\mathbf{W}_{\mathcal{P}} \leftarrow 0$\Comment{\emph{remove pruned neurons}}
            \State $\mathcal{R}=\mathcal{F}-\mathcal{P}$ \Comment{\emph{the remaining neurons}}
            \State Update $\mathbf{W}_{\mathcal{R}}$; $i \leftarrow i + 1$
        \EndWhile
    \end{algorithmic}
\end{algorithm}
Consider a neural network with $L$ layers, each layer  
specified by its weight $\mathbf{W}^{l}\in \mathbb{R}^{C_O^{l}\times C_I^{l}\times K^{l}\times K^{l}}$, $K$ being the kernel size and $C_I$ and $C_O$ being the number of input and output channels/neurons, respectively. Altogether, these parameters form the parameter set $\mathbf{W}=\{\mathbf{W}^{l}\}_{l=1}^L$ for the network. Given a training set consisting of $N$ input-output pairs $\{(x_i,y_i)\}_{i=1}^N$, learning the parameters of a network under filter sparsity constraints can be expressed as solving the following optimization problem:
\begin{equation}
    \argmin_{\mathbf{W}} \frac{1}{N} \sum_{i=1}^{N}\ell(y_i, f(x_i,\mathbf{W})) + r(\mathbf{W}),
 \quad \text{s.t.}\quad \frac{|\mathcal{P}|}{|\mathcal{F}|} \geq \alpha
\end{equation}
\noindent where $\ell(\cdot)$ denotes the loss function that compares the network prediction to the ground-truth, $f(\cdot)$ encodes the network transformation, $r(\mathbf{\cdot})$ is a regularizer acting on the network parameters, and $\alpha$ is the target pruning ratio. $\mathcal{F}=\{\mathcal{F}^{l}\}_{l=1}^L$ represents the index set of all the neurons in the network. This index set can be divided into two disjoints sets $\mathcal{P}=\{\mathcal{P}^{l}\}_{l=1}^L$ and $\mathcal{R}=\{\mathcal{R}^{l}\}_{l=1}^L$, representing the index sets of the pruned and remaining neurons respectively. We have $\mathcal{F}=\mathcal{P}\cup \mathcal{R}$ and $\mathcal{P}\cap \mathcal{R}=\emptyset$. PaT involves three stages: dense training, network pruning, and sparse training. During initial dense training, the forward/backward passes are computed using all the filters in the network $f(x_i,\textbf{W}_{\mathcal{F}})$. While during sparse training, the forward/backward passes only use the remaining neurons $f(x_i,\textbf{W}_{\mathcal{R}})$.
Akin to~\cite{molchanov2019importance,molchanov2016pruning}, we follow the paradigm of iterative pruning that finishes within one epoch. More precisely, as shown in \AlgRef{alg:iterative-pruning-algo}, the process consists of the following two steps. First, at each training iteration, we compute the importance metric of each neuron according to a pruning criterion. Meanwhile we keep updating network weights as normal.  
Then, at each pruning step, we get the averaged importance score for all neurons, and then remove the neurons with the smallest importance values. 
Each pruning step is carried out after seeing multiple training batches, usually several hundreds of batches are more than sufficient~\cite{molchanov2019importance}. Note that though containing several quick interactive steps, the entire pruning can be finished very quickly within one training epoch.

For comprehensiveness we consider two popular criteria from literature - both magnitude-based and gradient-based schemes for neuron importance ranking: 
\newline \noindent\textbf{Magnitude-based} criterion uses the $\mathnormal{l}_2$-norm of the neuron weights to measure the relevance of a neuron:  
\begin{equation}\label{mag-based-importance}
    \mathcal{I}_n^{l} = ||\mathbf{W}_n^{l}||_2/\sqrt{P^{l}},
\end{equation}
where $P^{l}=C_I^l\times K^l \times K^l$ denotes the number of parameters per filter at layer $l$, and $n$ specifies the output neuron index within $\mathbf{W}^l$. Such normalization ensures its comparability for neurons from different layers with different sizes~\cite{alvarez2016learning}. 
\newline \noindent\textbf{Gradient-based} criterion considers the Taylor expansion of the loss change to approximate the importance of a neuron. Initially, Molchanov~\etal \cite{molchanov2016pruning} proposed to estimate importance as a magnitude of the gradient-activation product, and more recently, SNIP~\cite{lee2018snip} and FORCE~\cite{de2020force} extended the idea to a parameter level. Specifically, gradient-based criterion using Taylor expansion for neurons can be defined as:
\begin{equation}\label{grad-based-importance}
    \mathcal{I}_n^{l} = \left|\sum_{w\in \mathbf{W}_n^{l}} g_{w}w\right|,
\end{equation}
\noindent where $g_w$ is the gradient of the weight $w$. The metric estimates an approximate change in the loss function once the neuron is removed. As suggested in~\cite{molchanov2019importance}, for networks using batch normalization, the best way to apply pruning is on the batch normalization layers instead of convolutional filters directly. Additionally, the loss of removing the channel can be approximated via accumulative effect of the learnable scale and shift: $\mathcal{I}_n^{l}=\left|g_{\gamma_n^{l}}\gamma_n^{l}+g_{\beta_n^{l}}\beta_n^{l}\right|$, where $\gamma$ and $\beta$ are the weight and bias of the batch normalization layer, respectively. We empirically observe slight improvements using $L_1$ for pruning during training rather than the original $L_2$ as in~\cite{molchanov2019importance} for post-training pruning.

\subsection{Towards Early Pruning}
\label{sec:policy}
Recall that our goal is to maximize the accuracy of the network while minimizing the compute required for training. This compute is usually dominated by the amount of time performing dense training. The sooner we prune the network, the less resources it requires to finish training.

As both prior work~\cite{Frankle2020earlyphase} and we empirically observe, the early stage of neural network training imposes a rapid motion in parameter space with large gradient magnitudes. This generates fruitful information for initial network convergence and a quick accuracy boost.
With such a fact, an intuitive option for early pruning can be to analyze the emergence of important neurons at different training stages that the network gradually picks up for the underlying task, and then prune the insignificant ones right away. 

Given its intrinsic access to weights and gradients, training allows one to quickly rank all the neurons globally with very little extra compute. This allows the network to take a quick glimpse into the problem from the architecture space that is empirically observed informative \cite{he2015deep,Howard2017MobileNetsEC,szegedy2015going}, while we can quite efficiently track architectural convergence. 

To this end, we check after each training epoch for a sub-network specified by the top $k$ most important neurons globally according to the chosen pruning criterion. Close to but different from a final winning ticket, an intermediate helps identify dominant neurons, but remains not as strong as its final version, while constantly changing during training. However, as we will show later, the architecture of such sub-networks changes rapidly in the first few epochs, then surprisingly shows minimal changes thereafter for the remaining epochs. Knowingly exploiting such fast convergence to stability and slow changes of dominate sub-network thereafter, we argue and demonstrate pruning can be started as early as when its \textit{Top-k} sub-network stabilizes. Next, we explore network similarity to signal such stability.

\begin{figure}[!t]
    \centering
    \includegraphics[scale=0.5]{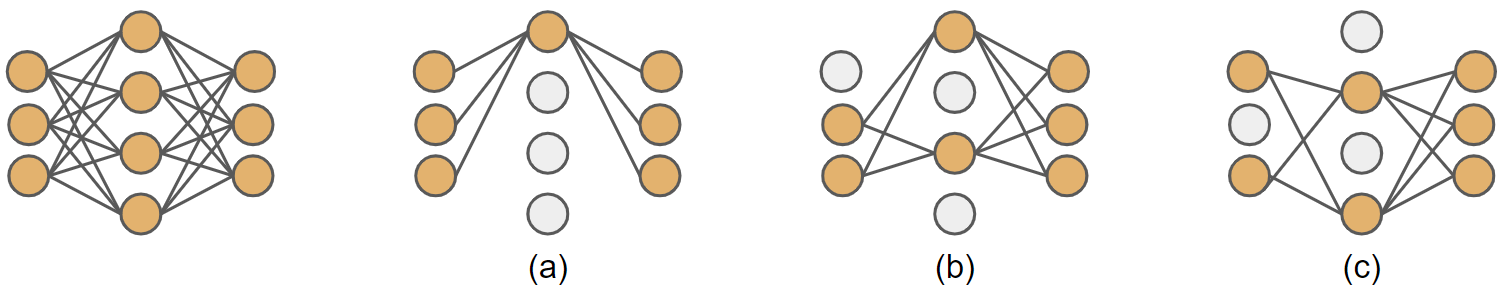}
    \caption{Structure of different sub-networks. Colored circles and solid lines are active neurons and connections. Sub-figures (a), (b) and (c) are three different sub-networks of the original network on the left. While these sub-networks having the same number of neurons in total, sub-networks (b) and (c) are in higher similarity.}
    \label{fig:subnet-structure-similarity}
    \vspace{-0.4cm}
\end{figure}

\begin{algorithm}[!t]
	\caption{Pruning-aware-Training (PaT)}
	\label{algo:stability-guided-Pat-algo}
	\textbf{Input:} Network with random initialized weights $\mathbf{W}_{\mathcal{F}, 0}$, stability threshold $\tau$, pruning ratio $\alpha$, total epochs $T$ as in original recipe\\
	\textbf{Output:} Pruned structure $\mathcal{R}$; trained weights $\mathbf{W}_{\mathcal{R}, T}$
	\begin{algorithmic}[1]
	    \State $\text{enforce epoch status} \in \{\text{dense}, \text{prune}, \text{sparse}\}$
	    \State epoch status $\leftarrow$ dense
	    \For{epoch $t=0,1,\ldots,T$}
	        \If{ epoch status \textbf{is} dense}
	            \State Train $\textbf{W}_{\mathcal{F}, t}$ by gradient descent
	            \State Get importance score averaged over the epoch
	            \State Get $\mathcal{N}_t$
	            \State Get $\text{EPI}_t$ with Eq.~\eqref{eq:epi_equation}
	            \If{$(\text{EPI}_t \geq \tau)$ and $(\text{EPI}_t \geq \text{EPI}_{t-j})_{1\leq j\leq 5}$} 
	            \State epoch status $\leftarrow$ prune
	            \EndIf
	       \ElsIf{epoch status \textbf{is} prune}
	            \State Prune $\alpha |\mathcal{F}|$ neurons with Algorithm \ref{alg:iterative-pruning-algo}
	            \State Get $\mathcal{P}$, update $\mathcal{R}$
	            \State epoch status $\leftarrow$ sparse
	       \Else
	            \State Train $\textbf{W}_{\mathcal{R}, t}$ by gradient descent
	       \EndIf
	    \EndFor
	    \State Return $\mathcal{R}$, $\mathbf{W}_{\mathcal{R},T}$
	\end{algorithmic} 
\end{algorithm}

\subsection{Early Pruning Indicator (EPI)}
\label{stability-intro}
We look into the structural similarity between dominant sub-networks to quantify architectural changes during training. Under a global neuron pruning scheme, merely using pruning ratio as a guidance fells short for this task: each pruning ratio can be easily satisfied by multiple variants, each sharing the same number of neurons while differing in architectures (see examples in Fig.~\ref{fig:subnet-structure-similarity}). As an alternative, we examine the distribution of the number of remaining neurons across all layers per pruned network.

Consider two sub-networks $\mathcal{N}_1$ and $\mathcal{N}_2$ under the same prune ratio containing the same number of remaining neurons. Let $n_{(1,l)}$ and $n_{(2, l)}$ be the number of neurons of $l_{th}$ layer in nets $\mathcal{N}_1$ and $\mathcal{N}_2$ respectively, then set $\{n_{(1, 1)}, n_{(1, 2)}, \cdots, n_{(1, L)}\}$ describes the structure of the sub-network $\mathcal{N}_1$, and similarly for $\mathcal{N}_2$. For the $l_{th}$ layer, we define the normalized difference between $\mathcal{N}_1$ and $\mathcal{N}_2$ as
\begin{equation}
\label{layer_distance}
    d_l(\mathcal{N}_1, \mathcal{N}_2) = \frac{\lvert n_{(1,l)}-n_{(2,l)}\rvert}{n_{(1, l)}+n_{(2,l)}},
\end{equation}
yielding a range from zero to one. The lower the distance, the closer the layer structure is. On top of this we can now construct a pruning stability indicator $\Psi$ combining the similarity for all the layers in the network:
\begin{equation}
    \Psi(\mathcal{N}_1, \mathcal{N}_2) = 1 - \frac{1}{L}\sum_{l=1}^{L}d_l(\mathcal{N}_1, \mathcal{N}_2),
\end{equation}
\noindent where $\Psi$ ranges from $0$ to $1$, with a lower value indicates high variations between the two sub-networks, and a high value indicates stability in the resulting network structure.

Given a pruning stability indicator, the algorithm to decide when to prune is described in Algorithm~\ref{algo:stability-guided-Pat-algo}. We first calculate the neurons' importance scores according to the pruning criterion at the end of each epoch $t$. Get the top $k$ neurons by ranking the importance scores and the structure indicator $\mathcal{N}_t=\{n_{(t,1)}, n_{(t,2)}, \cdots, n_{(t, L)}\}$ where $\sum_{l=1}^L n_{(t,l)} = k$ and $n_{t,l}$ is the number of neurons in the $l_{th}$ layer. Then calculate the sub-network structure similarities between $\mathcal{N}_t$ and $\mathcal{N}_{t-j}$ for $1\leq j\leq r$ where $r$ is the range of past epochs that we want to have a structure comparison.  We use the averaged structure similarity to reflect the structure stability, namely:
\begin{equation}
    \text{EPI}_t=\frac{1}{r}\sum_{j=1}^{r}\Psi (\mathcal{N}_t, \mathcal{N}_{t-j}).
\label{eq:epi_equation}
\end{equation}
This structure stability score is constantly increasing during training. When it reaches a certain threshold $\tau$, we can safely say that the resultant sub-network is reliable to achieve a good performance and we can start the pruning.

\section{Experiments}

We next experiment with varying architectures and pruning methods to showcase the strength of our proposed method for the classification task. In the appendix, we also demonstrate applicability to the task of object detection.

\noindent
\textbf{Experimental Settings.}
We prune ResNet34, ResNet50 and MobileNetV1 neural network architectures on the ImageNet ILSVRC 2012 dataset~\cite{imagenet} ($1.3$M images, $1000$ classes). Unless otherwise stated each pruning uses one single node with $8$ NVIDIA Tesla V100 GPUs. All experiments share the original training pipeline following PyTorch mixed-precision training under NVIDIA's recipe~\cite{nvidiarecipe} with $90$ epochs in total. The learning rate is warmed up linearly in the first $8$ epochs, then follows a cosine decay over the entire training. We use PyTorch Distributed Data Parallel training and for each GPU with an individual batch size at $128$. Our unpruned models achieve $77.32\%$ top-1 accuracy with ResNet50, $74.36\%$ with ResNet34 and $72.93\%$ with MobileNetV1.

\subsection{Understanding Early Pruning Epochs}

\begin{figure}[!t]
    \centering
    \vspace{-0.4cm}\includegraphics[width=0.6\columnwidth]{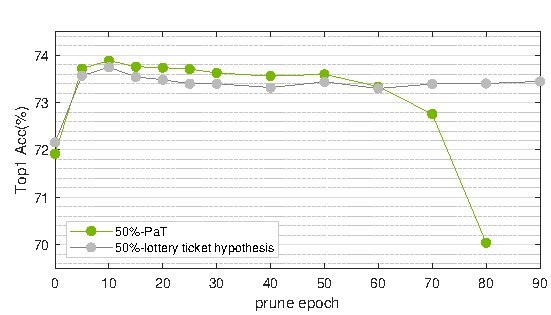}
    \caption{The performance of the PaT algorithm vs. lottery ticket hypothesis~\cite{frankle2019lottery} on ResNet50-ImageNet, the dense version achieves $77.32\%$ accuracy. \textbf{Green line:} the accuracy of PaT (jointly pruning and training) pruning at epoch $x$ with $90$-epoch training in total; \textbf{Gray line:} the network accuracy by applying the corresponding mask in PaT to initialization and train the sub-network from scratch, as the lottery ticket hypothesis~\cite{frankle2019lottery} does.}
    \label{fig:apply-mask-to-init}
     \vspace{-0.4cm}
\end{figure}

We start with understanding in depth the variations in final accuracy as a function of the starting pruning epoch. To do so, we analyze the accuracy changes by varying the starting pruning epoch, and continue training to the final epoch and check the associated accuracies. 

\noindent
\textbf{Pruning at different epochs.} \figref{fig:apply-mask-to-init}, in green, shows the top-1 accuracy obtained by pruning $50\%$ of the neurons on a ResNet50 using gradient-based criterion at various epochs during the $90$-epoch training cycle. 
As shown, the accuracy drop for late pruning is significant as there is not enough time left for recovering. We also observe that, compared to pruning at initialization (at epoch $0$), pruning after a few epochs consistently yields better performance. However, for all these experiments there is always a certain accuracy drop compared to the unpruned upper bound ($77.32\%$).

\noindent
\textbf{Lottery ticket hypothesis for structural pruning.} To better understand the role of early training with a dense model rather than a pruned model, we evaluate the idea of lottery-ticket hypothesis for structural pruning. We follow~\cite{frankle2019lottery} and train from scratch a sub-network obtained by pruning using the original initialization. All pruning masks are collected from the previous experiment (Fig.~\ref{fig:apply-mask-to-init}). Results are shown in the same plot with a gray line. Note that, due to iterative nature of the pruning algorithm, for pruning at $0$ we use the mask when the epoch is finished. When it is applied at the initialization as a lottery ticket, the final accuracy is slightly different. From these results, we can conclude that, in the structural pruning case, the lottery ticket hypothesis may not hold. Pruning the network during training performs better than training a winning ticket in isolation from scratch.

\begin{figure*}[!t]
  \centering
  \begin{tabular}{c}
  \includegraphics[width=0.5\columnwidth]{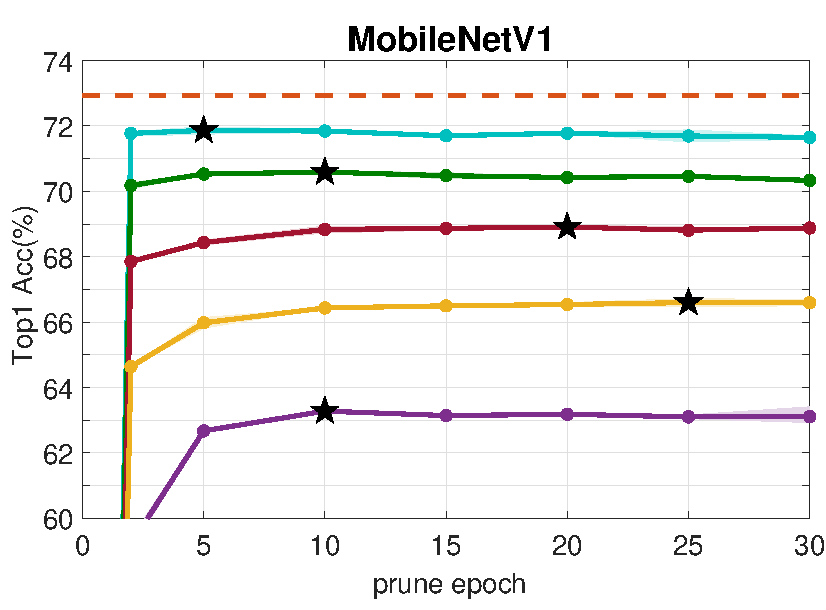}
  \includegraphics[width=0.5\columnwidth]{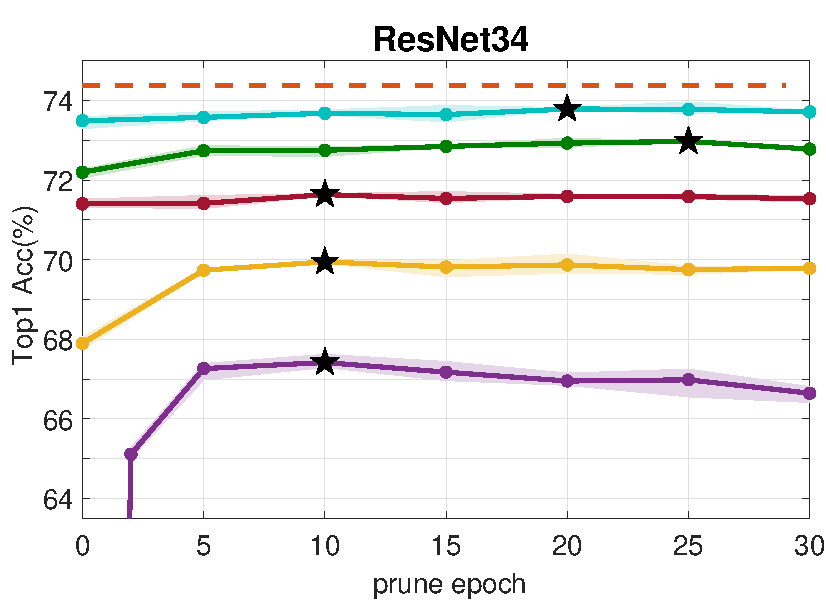} \includegraphics[width=0.5\columnwidth]{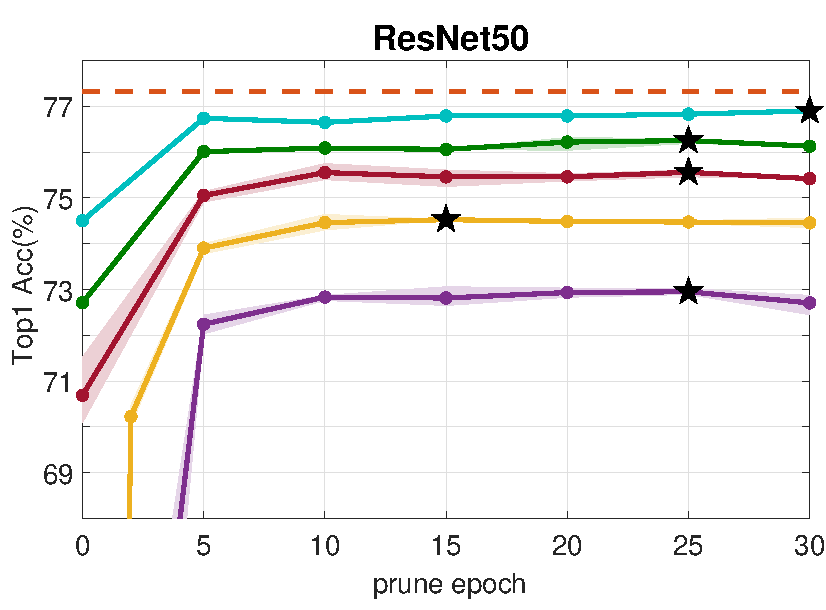}
  \includegraphics[width=0.5\columnwidth]{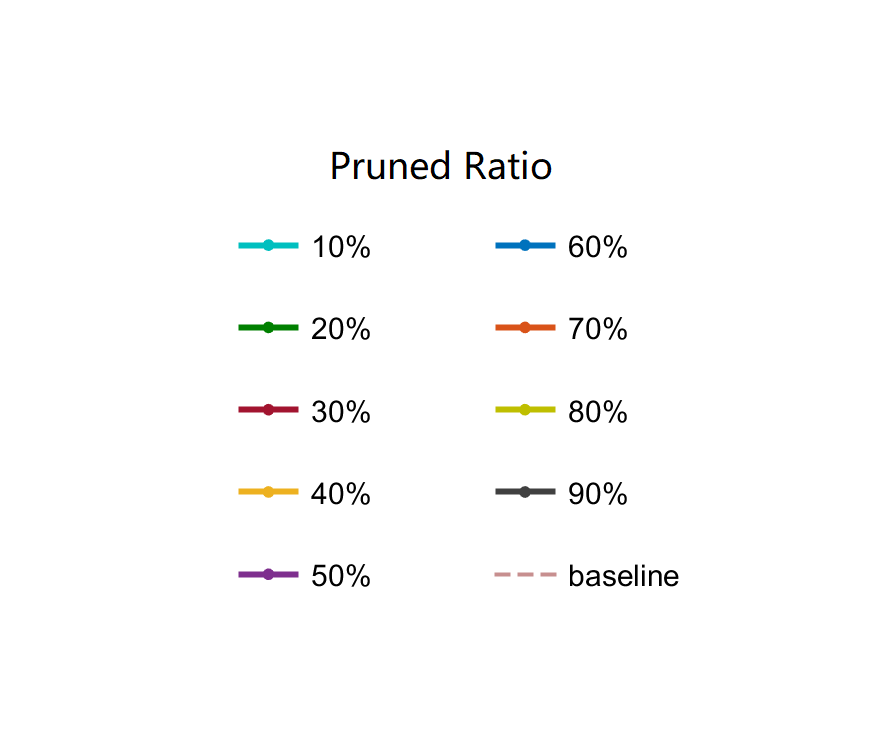}\\
  (a) \small{Magnitude-based pruning} \\
  \includegraphics[width=0.5\columnwidth]{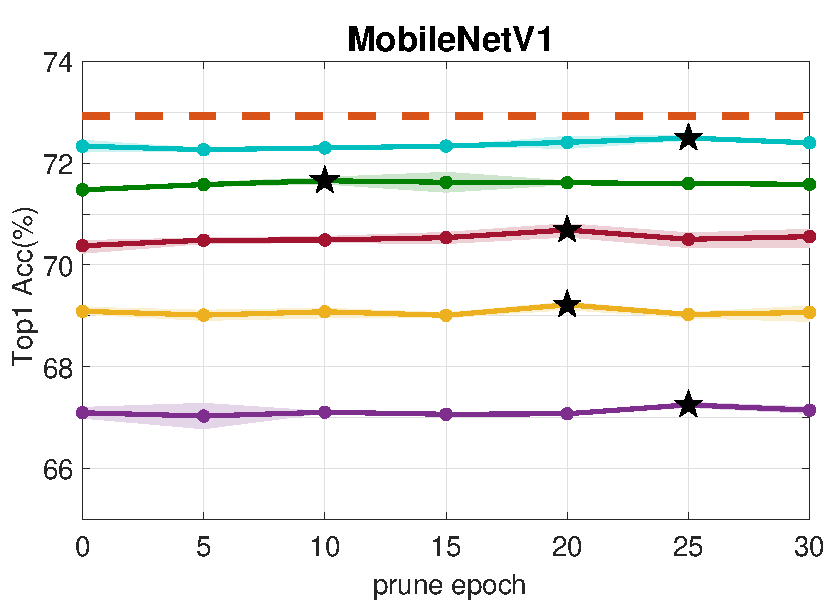}
  \includegraphics[width=0.5\columnwidth]{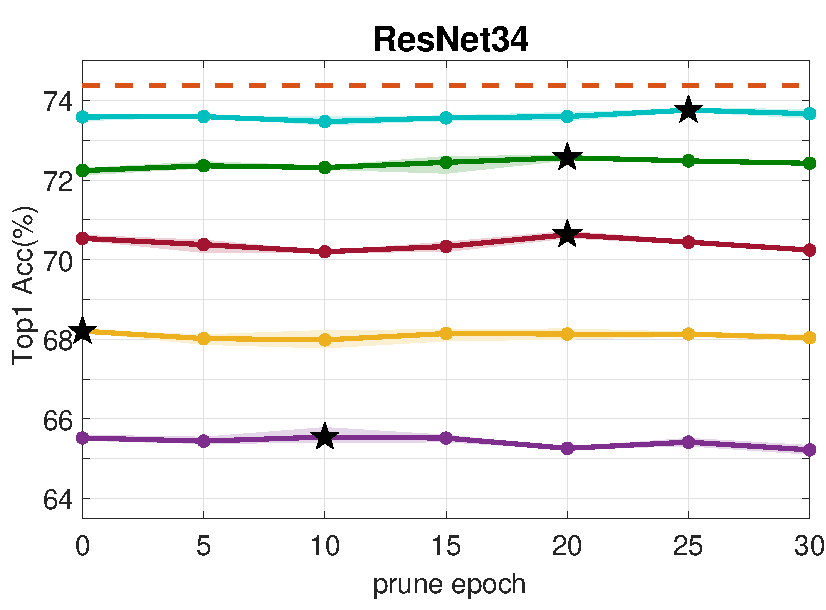}
  \includegraphics[width=0.5\columnwidth]{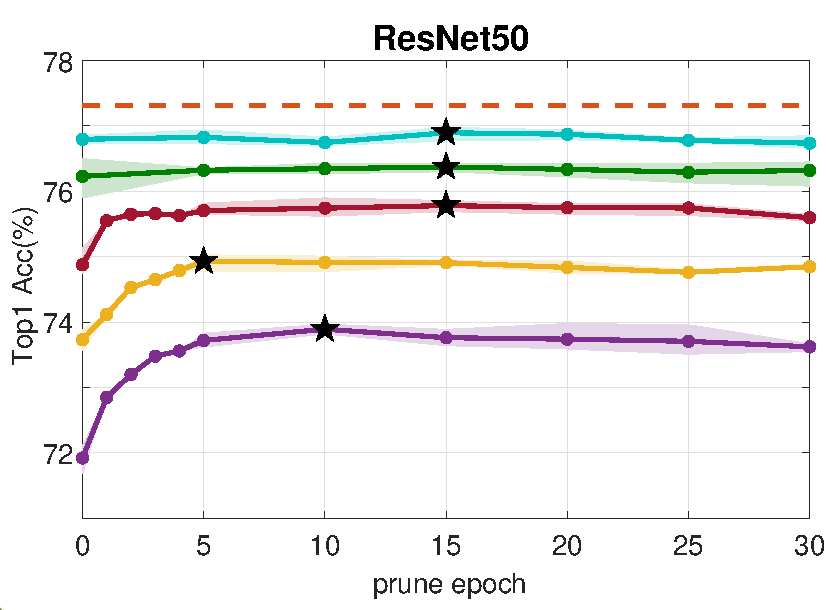}
  \includegraphics[width=0.5\columnwidth]{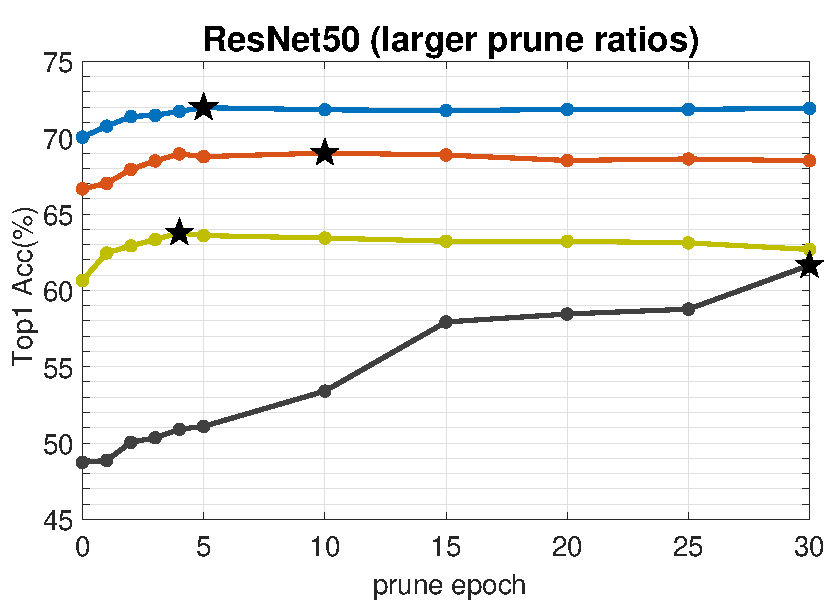}\\
    (b) \small{Gradient-based pruning} 
  \end{tabular}
    \caption{The final ImageNet Top-1 accuracy of the pruned network when pruning occurs at different epochs during the early stage of training. We observe pruning at initialization tends to result in untrainable network with magnitude-based pruning method. For gradient-based method, we observe a higher degradation occur when more filters are pruned, and show pruning ratios up to $90\%$ on ResNet50. [Prune Ratio] denotes the percentage of neurons removed.}
    \vspace{-4mm}
    \label{fig:prune-result}
\end{figure*}

\noindent
\textbf{Varying pruning ratio and architecture.} We now take a closer look at the accuracy drop incurred when pruning occurs during the early stage of training (first $30$ epochs from Fig.~\ref{fig:apply-mask-to-init}). Fig.~\ref{fig:prune-result} shows these results for different architectures using magnitude-based and gradient-based pruning respectively. As we can see, for magnitude-based pruning, there is a significant drop in accuracy if the pruning occurs too early, especially for large prune ratios. This effect is particularly clear if, for instance, we prune $50\%$ neurons of a ResNet50 at epoch $0$ which makes the network not trainable. This is expected as the pruning ratio is large and the weights have not been updated at all. Therefore, it is not possible to estimate the importance of each neuron correctly. For gradient-based pruning, the accuracy drop varies depending on the architecture. In this case, pruning at initialization has less impact compared to magnitude-based pruning. For instance, pruning a ResNet34 or a MobileNetV1 leads to minimal drop in accuracy. For ResNet50, however, the accuracy drop increases as the pruning ratio increases. Thus, late pruning would be preferred to maximize performance. Let's see how we can find the optimal pruning epoch during a single training session. 

\begin{figure}[!t]
\centering
\resizebox{0.8\columnwidth}{!}{
\renewcommand*{\arraystretch}{0.3}
\begin{tabular}{c}
\includegraphics[width=\columnwidth]{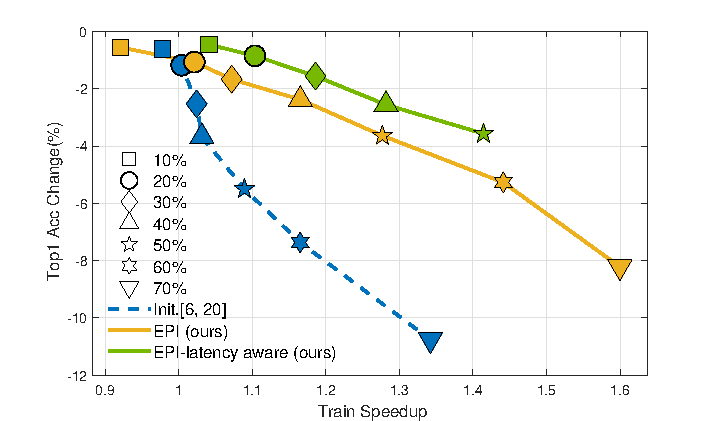}\\
\end{tabular}
}    
\caption{Actual training speed-ups on ResNet50 with different prune ratios using different pruning policies. Actual speed measured on an NVIDIA TITAN V GPU at batch size $64$. Top-right corner is preferred.}  \vspace{-4mm}
    \label{fig:policy-compare}
\end{figure}

\label{sec:epi-guided-pruning-result}
\begin{table}[!t]
    \centering
    \begin{minipage}{\columnwidth}
    \resizebox{.99\columnwidth}{!}{
    \begin{tabular}{cccccc}
        \toprule
        \multicolumn{1}{c}{}& \multirow{2}{*}{\textbf{Network(s)}} & \multicolumn{4}{c}{\textbf{Starting epoch for pruning}} \\  \cline{3-6}
        
        \multicolumn{1}{c}{}&  & Random & Init.~\cite{de2020force, lee2018snip} & Pre-defined~\cite{Frankle2020earlyphase}  &  EPI (ours) \\
        \midrule
        \multirow{4}{*}{\begin{sideways}\textbf{\footnotesize{Magnitude}}\end{sideways}  } & ResNet50 & $0.940$ & $3.604$ & $0.115$ & $\mathbf{0.091}$ \\
        & ResNet34 & $0.267$ & $0.838$ & $0.353$ & $\mathbf{0.169}$ \\
        & MobileNet-v1 & $6.285$ & -- & $0.135$ & $\mathbf{0.135}$ \\    
        \cline{2-6}
        & Overall & $2.497$ & $2.221$ & $0.201$ & $\mathbf{0.132}$ \\
        \midrule
        \multirow{4}{*}{\begin{sideways}\textbf{\footnotesize{Gradient}}\end{sideways}  } & ResNet50 & $0.992$  & $2.738$ & $0.267$ & $\mathbf{0.092}$ \\
        & ResNet34 & $0.153$ & $\mathbf{0.122}$ & $0.221$ & $0.195$ \\
        & MobileNet-v1 & $0.132$ & $0.186$ & $\mathbf{0.110}$ & $0.178$ \\
        \cline{2-6}
        & Overall & $0.426$ & $1.015$ & $0.199$ & $\mathbf{0.155}$ \\
        \bottomrule
    \end{tabular}
    }
    \end{minipage}
    \caption{Absolute top-1 accuracy change relative to the results per oracle grid search for optimal pruning starting point. Lower is better. Init. refers to pruning at initialization using a structural version of FORCE~\cite{de2020force}, also equivalent to the structural and iterative SNIP~\cite{lee2018snip}. Pre-define refers to heuristically pre-define a pruning start epoch $30$, which we find empirically the last epoch not leading to a significant accuracy drop. This is aligned with~\cite{frankle2019linear} suggesting waiting for several training iterations before pruning.}
     \vspace{-0.4cm}
    \label{tab:policy-compare}
\end{table}

\subsection{EPI-guided Pruning}
Given the previous results, we now demonstrate the ability of our approach to determine the optimal pruning epoch. Thus, in this experiment we compare our policy to a heuristic and a random policies. For heuristic policy, we consider setting the pruning epoch to $0$ which is equivalent to pruning at initialization~\cite{de2020force,lee2018snip}. For the random policy, we select randomly a pruning epoch during the early stage of training, \ie in the range $[0, 30]$. We repeat the random policy experiment $100$ times and report the mean of the results.

\noindent
\textbf{Selecting the EPI threshold ($\tau$).} Our method introduces a hyperparameter $\tau$ such that when EPI (Eq.~\ref{eq:epi_equation}) reaches it we can start pruning. We find that a universal value can be used for all architectures and all pruning ratios, however, it is sensitive to the pruning algorithm. To this end, we perform a sensitivity analysis on ResNet50 over pruning ratios of $10\%$--$50\%$ with increments of $10\%$ and use a grid search to set the value that yields the best pruning result. As a result, we find $\tau= 0.983$ for magnitude-based pruning and $\tau= 0.944$ for gradient-based pruning to be the best. We tuned this value for ResNet50, however, we will show its generalizability by performing tests on ResNet34, MobileNetV1 in the main text and SSD in the Appendix. 

\noindent
\textbf{Policy comparison.}
\tabref{tab:policy-compare} shows the results of experiments for importance and magnitude based pruning under the guidance of our proposed EPI and the universal EPI threshold $\tau$. We compare the results with random policy and heuristic policy of pruning at initialization. 
We report the average accuracy drop compared to best accuracy achieved via grid search for each network. The values for ResNet50 gradient-based pruning is averaged over prune ratios $10\%$--$90\%$; all the rests are averaged over prune ratios $10\%$--$50\%$. We also report the overall average accuracy drop over three networks when using different policies to guide the pruning in row ``overall''.
As shown, our policy clearly yields a significantly higher performance compared to random pruning. In the case of gradient-based pruning, our approach performs on par compared to heuristics on ResNet34 and MobileNetV1. Overall our approach performs better with less top-1 accuracy change. For magnitude-based pruning, our approach yields significantly better results compared to heuristics.
The optimal pruning epoch varies for different architectures and different prune ratios, see the appendix for the sensitivity analysis to the stability threshold.

\begin{figure*}[!t]
\vspace{-0.4cm}
   \centering
   \begin{tabular}{cc}
   \includegraphics[width=0.5\columnwidth]{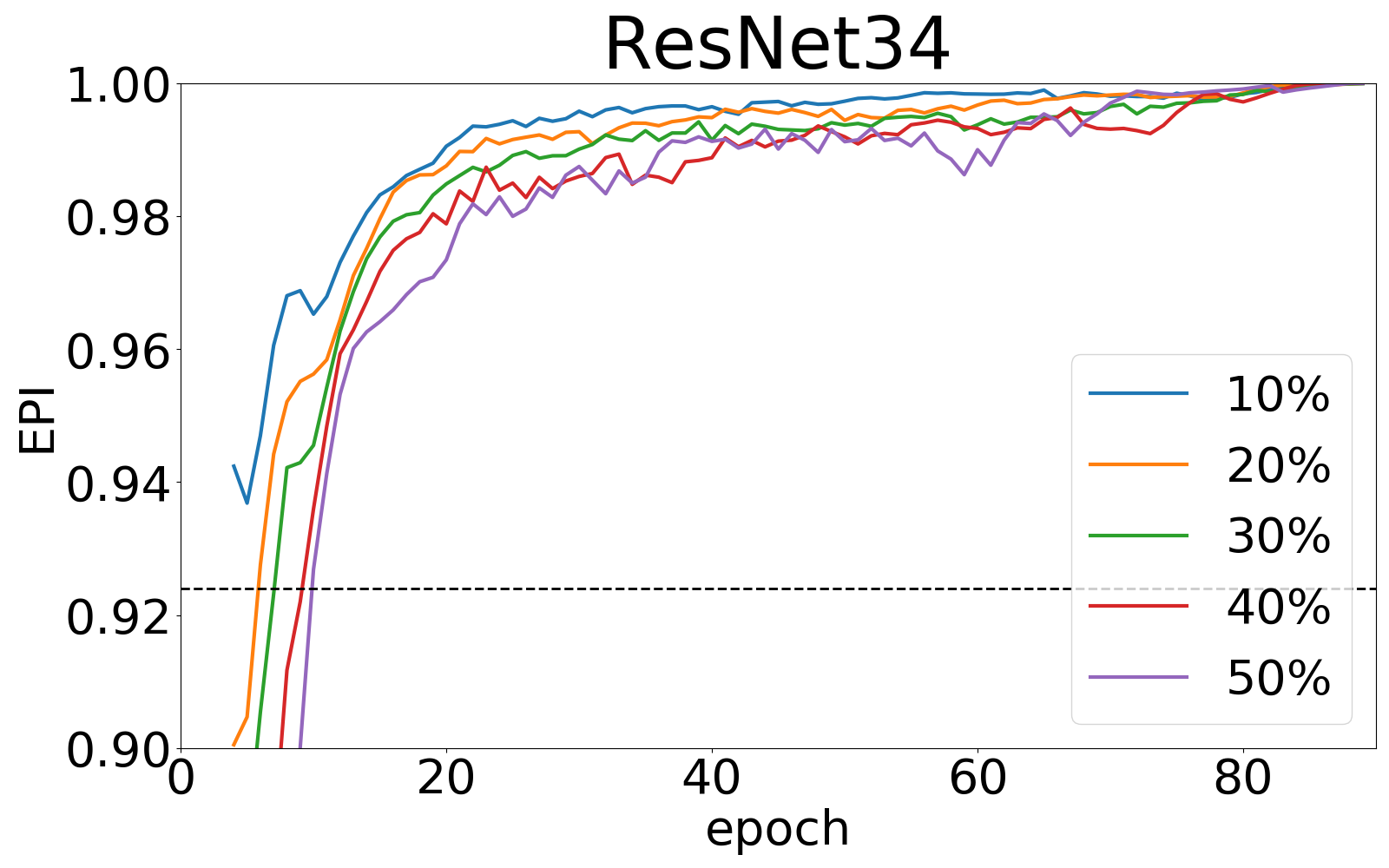} \includegraphics[width=0.5\columnwidth]{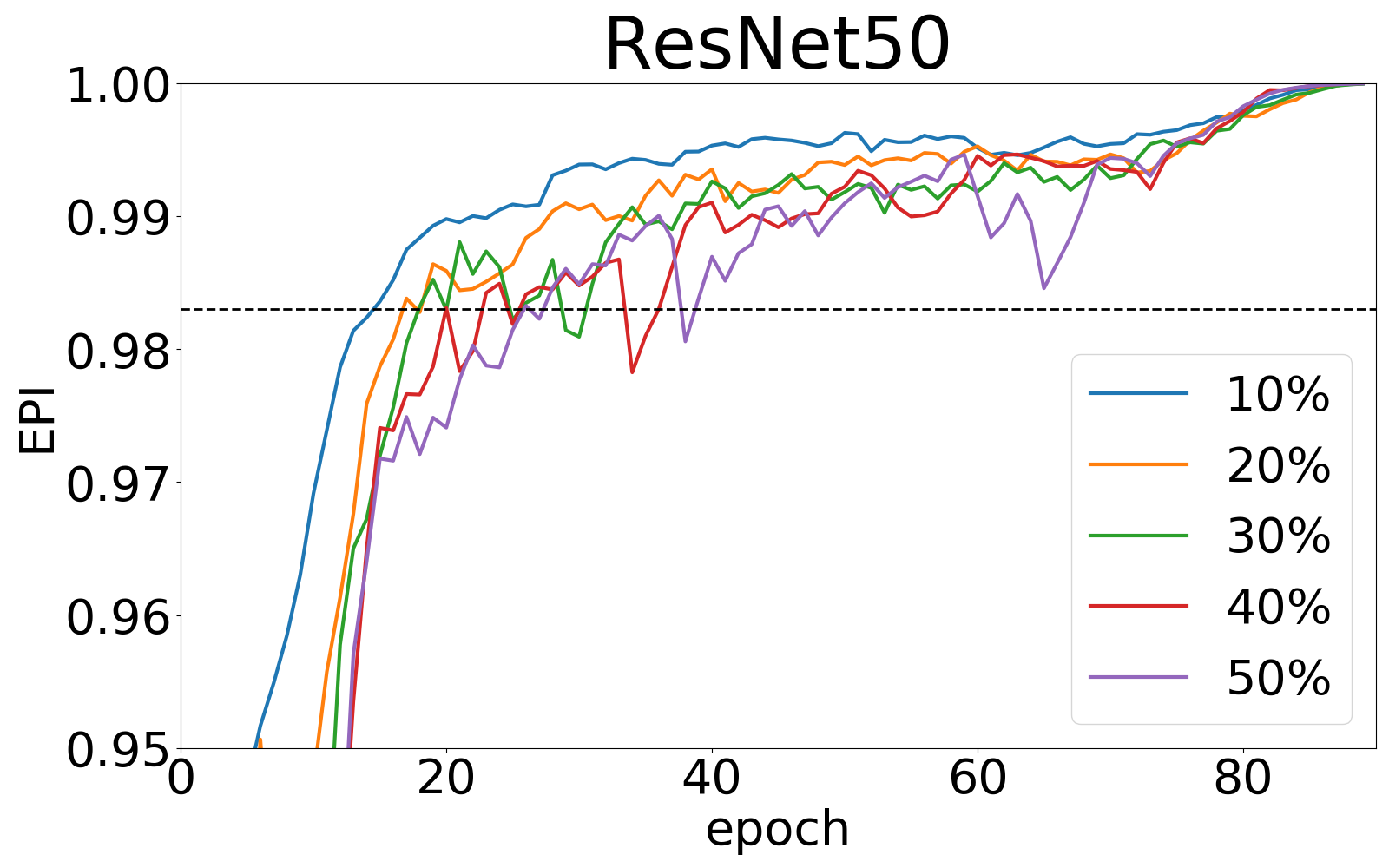}&
  \includegraphics[width=0.48\columnwidth]{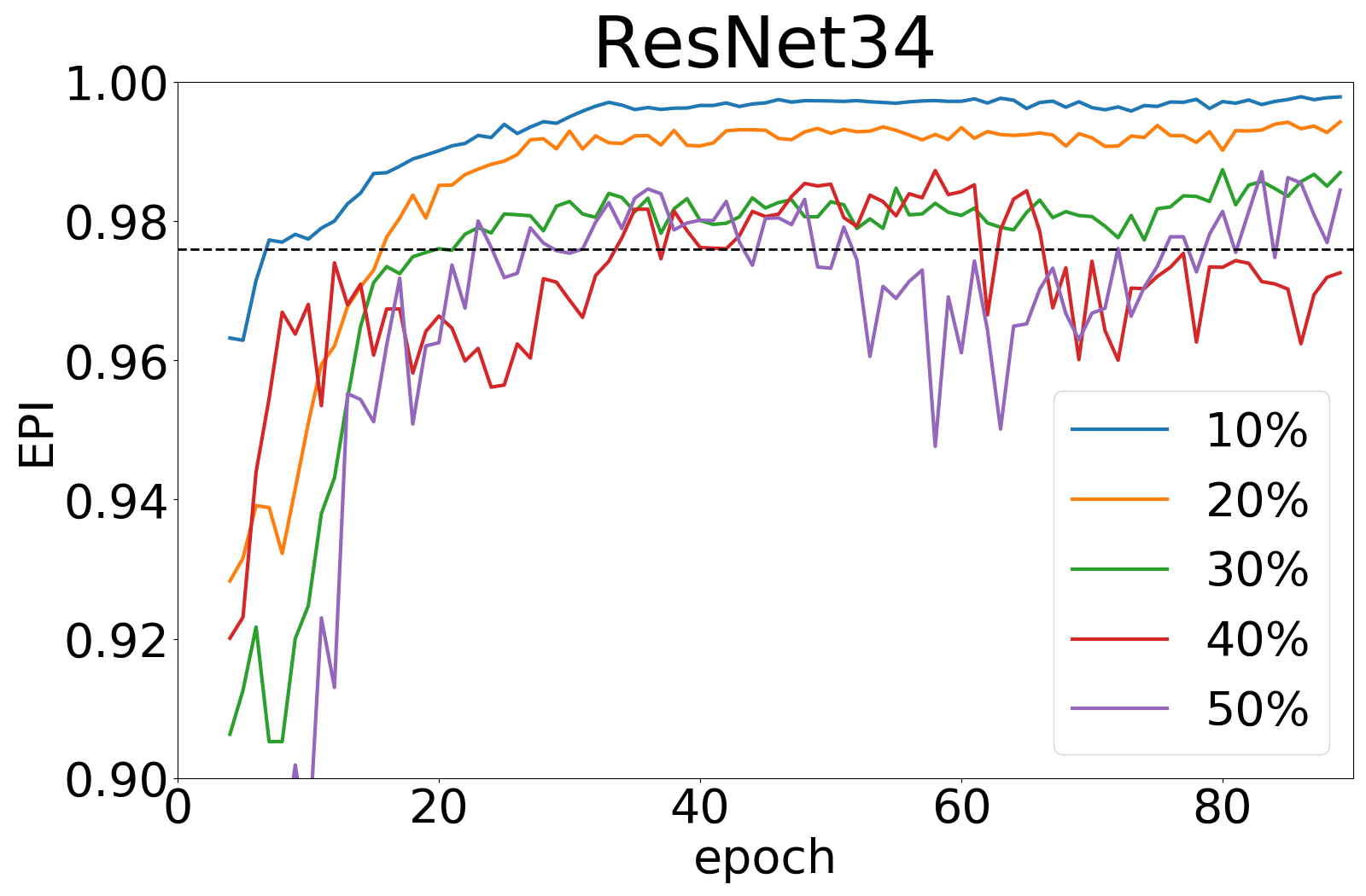}
  \includegraphics[width=0.5\columnwidth]{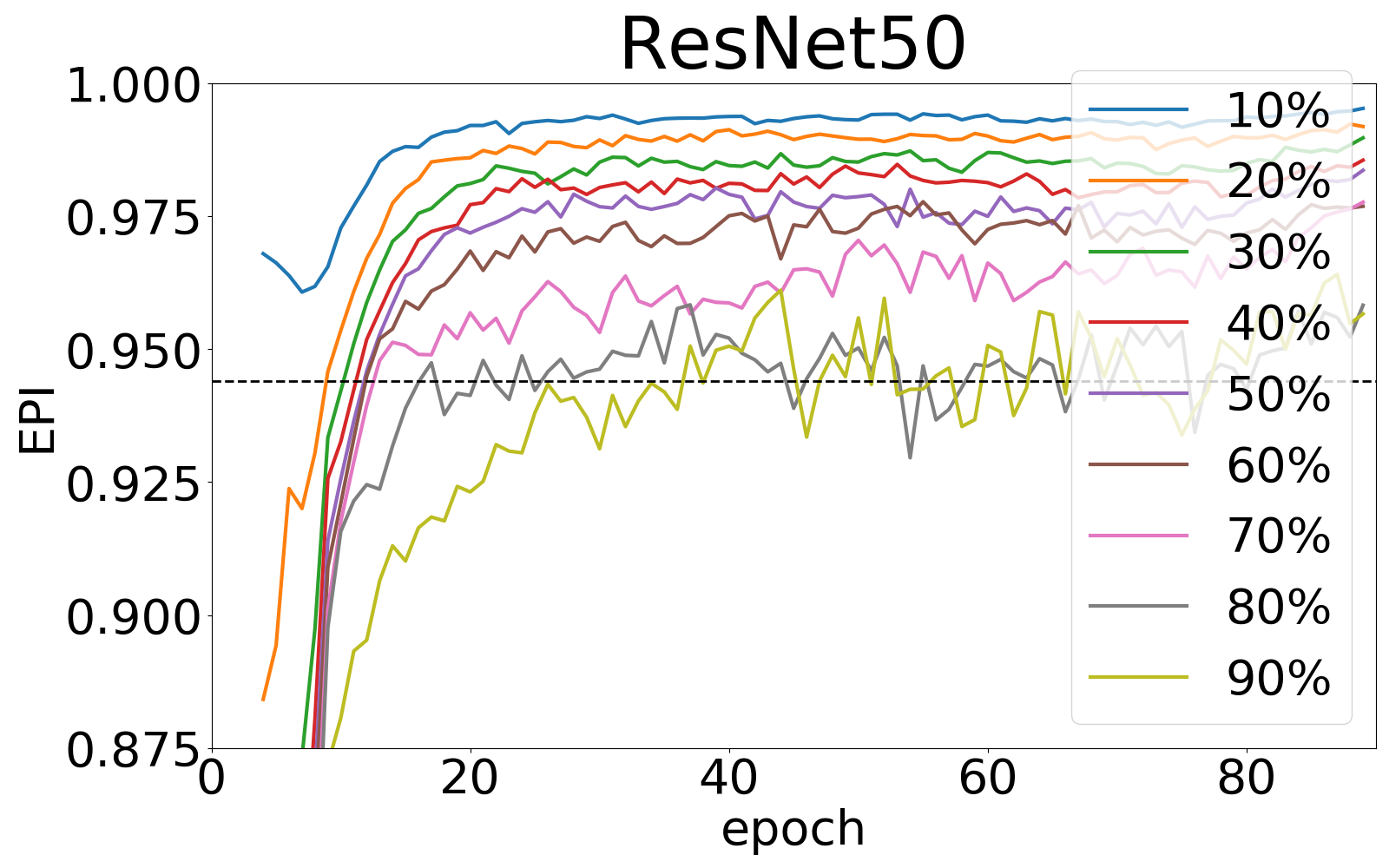}\\
   (a) magnitude-based EPI &(b) gradient-based EPI
   \end{tabular}
    \caption{Structure stability analysis for ResNet architectures for magnitude-based (a) and gradient-based (b) pruning. Dashed line shows the EPI threshold selected for each network under the pruning criterion. Results for MobilnetV1 can be found in the supplemental material.}
    \label{fig:topk-structure-stability}
    \vspace{-0.4cm}
\end{figure*}

\subsection{Training Speedup} 

We also calculate the actual training speedup when using our proposed EPI~(Eq.~\ref{eq:epi_equation}) policy and the heuristic ones on ResNet50 gradient-based pruning. Fig.~\ref{fig:policy-compare} shows the policies comparison result. 

\noindent
\textbf{Comparing  to pruning at initialization.}
When compared with pruning at initialization (heuristic pruning at epoch $0$), we achieve larger speed-up although we start prune later. 
That happens because of structural pruning, where pruning at different epochs might result in different structures: faster or slower. It turns out that pruning at epoch $0$ is very inefficient, as most neurons pruned are from deeper layers resulting in slow models; therefore, training speed-up is not huge and training such a model is more expensive than the one resulted from pruning at a later epoch. In the meantime, pruning at zero leads to larger accuracy drop especially with large prune ratios.

\noindent
\textbf{Latency-aware pruning.} We further apply latency-aware pruning using our policy, which aims to reduce the latency of the model and not only the number of parameters. For that, we penalize the neuron group's saliency with the latency reduction resulting from pruning them. Those neurons requiring larger compute will have lower importance and, therefore, more likely to be pruned. As shown in Fig.~\ref{fig:policy-compare}, using EPI-latency aware pruning yields models that are more GPU-friendly and faster.

\noindent
\textbf{Training cost comparison.} Training a dense ResNet50 on ImageNet with $8$ NVIDIA Tesla V100 GPUs takes around $9$ hours and costs around $\$ 220$ on AWS. Considering a $50\%$ pruning ratio, the training cost of our method is $\$154$ as we achieve up to $30\%$ training time reduction. In contrast, the total training cost for train-prune-finetune methods is around $\$364$ as they need $90$ additional retraining epochs. Thus, a $2.4\times$ training cost reduction. 

\begin{table}[!t]
    \centering
    \resizebox{\columnwidth}{!}{
        \begin{tabular}{cccccc}
        \toprule
        & \textbf{Method} & \textbf{Top-1 acc.} $\uparrow$ & \textbf{FLOPs (G)}$\downarrow$ & \textbf{FLOPs reduc.(\%)} $\uparrow$ & \textbf{Starting epoch} $\downarrow$ \\
        \midrule
        
        \multirow{2}{*}{\textbf{ResNet34}} & GPWP~\cite{GPWP} & $\mathbf{73.64}$ & $3.087$ & $15.90$ & $40$\\
        & PaT \textbf{(ours)} & $73.50$ & $\mathbf{2.911}$ & $\mathbf{20.70}$ & $\mathbf{11}$ \\
        \midrule
        \multirow{3}{*}{\textbf{ResNet50}} & PT~\cite{lym2019prunetrain} & $74.73$ & $2.303$ & $44.00$ & $90$ \\
        & LFPC~\cite{he2020learning} & $74.18$  & $\mathbf{1.612}$ & $\mathbf{60.80}$ & $35$\\
        & PaT \textbf{(ours)} & $\mathbf{74.85}$ & $1.695$ & $58.78$ & \textbf{13}\\
        \hline
        
    \end{tabular}
    }
    \footnotesize
    \caption{Comparison with state-of-the-art in-training pruning methods. For a fair comparison we report here $10\%$ filter pruning for ResNet34 and $40\%$ filter pruning for ResNet50 as literature.
    }
    \vspace{-0.5cm}
    \label{tab:prune-during-training-compare}
\end{table}

\subsection{Comparisons with the State-of-the-art} 
We also compare our method to prior arts on ImageNet dataset and present results in Tab.~\ref{tab:prune-during-training-compare}. Compared to GPWP~\cite{GPWP}, our method yields lower accuracy but a higher reduction in FLOPs. Compared to PruneTrain~\cite{lym2019prunetrain}, we achieve significantly higher accuracy and a larger reduction in FLOPs. LFPC ~\cite{he2020learning} reduces $2\%$ more FLOPs, but the accuracy loss is $0.67\%$ larger. Importantly, LFPC requires longer training time as it needs additional epochs to train a sampler. Overall, we prune much earlier than other techniques, and, therefore, reduce more training time. Moreover, while other techniques use heuristics to define the pruning epoch, we propose an automatic metric that scales with a universal threshold to automatically determine an early point. This helps create a new benchmark protocol by evaluating prior methods under the same setting.

\subsection{Ablation Studies}
\label{sec:ablation}

\noindent \textbf{Stability analysis.}
The goal of this experiment is to demonstrate that during the early stage of training, the dominate sub-network architecture varies significantly and slowly converges to the final architecture as the training progresses. To this end, we train different architectures to convergence and, in the process, we compute their EPI scores for sub-networks under different pruning ratios. Fig.~\ref{fig:topk-structure-stability} shows the results for this experiment for magnitude-based and gradient-based criterion respectively. As shown, the EPI score decreases as the pruning ratio increases. We also observe that, independently of the pruning method used, the stability grows rapidly in the early stage of training and then continues increasing steadily for later stages. These results are consistent with those presented in~\cite{Frankle2020earlyphase} showing a significant change in the network architecture at the beginning but not towards the end of training. 

\noindent\textbf{Performance of similar structures.}
The goal of this experiment is to provide empirical support to the assumption that sub-networks with similar structures will likely perform similarly. That is, the performance relies on the structure rather than the neurons being selected during pruning. We use the ResNet50 and obtain the neuron pruning mask at epoch $10$ using $50\%$ pruning ratio using gradient-based ranking. Next, we obtain $10$ variations of this mask by selecting different neurons in each layer while maintaining the number of neurons per layer. As a result, we get different neuron masks but the same sub-network structure. We train the initially pruned network to convergence and obtain a top1 accuracy of $73.98\%$. Finally, we evaluate the performance of training to completion the 10th checkpoint pruned using the $10$ mask variations yielding an average top1 accuracy difference of only $0.36\% \pm 0.13\%$. 
We also randomly generate $5$ variations of masks that lead to different sub-network structures, which have an around $0.8$ similarity score to the structure resulted from the original mask. We get an average performance difference of $0.53\% \pm 0.39\%$.
Note that, for a fair comparison, we use the same checkpoint to minimize randomness due to different initialization. From these results, we can conclude networks with the same structure perform similarly. With different structures, the performance varies more. 

\noindent\textbf{Compared to pruning pre-trained models.} 
In many practical scenarios, people train a model (or use a pre-trained backbone) to solve the task at hand. However, there are computation or latency constraints at deployment. People might want to train a smaller model from scratch or prune and finetune the existing model. Both scenarios would benefit from EPI, as we show next.

We compare pruning during training versus pruning pre-trained models in  Tab.~\ref{tab:pruning-compare}. 
For pruning a pre-trained model (loading Pytorch weights from ResNet50 trained for $90$ epochs, with acc. $77.32\%$), we do pruning in the same setting and then finetune it for $90$ epochs (marked as \textit{post trained pruning}). For a fair comparison when we apply EPI on a model trained from scratch we increase the total number of epochs to $90+90=180$ (column \textit{EPI, equal epochs}). We clearly outperform pruning pre-trained weights, while saving compute costs. We get pruned models earlier, making the training even faster.
Models from EPI can be finetuned for the same amount of clock time as \textit{post trained pruning}, shown as column \textit{EPI, equal training time}. This improves results even more as models get trained for more epochs. Again, applying EPI speeds up training right after pruning is complete, while pruning on pre-trained benefits pruning only after the training is finished. 

\begin{table}[t]
    \begin{minipage}{\columnwidth}
    \resizebox{.99\columnwidth}{!}{
    \begin{tabular}{|c | c c c |}
         \hline
         prune ratio & post trained pruning & EPI, equal epochs & EPI, equal training time \\
         \hline
         $10\%$ & $77.17\%$ & $\mathbf{77.69\%}$ & $\mathbf{77.74\%}$ \\
         $20\%$ & $76.78\%$ & $\mathbf{76.82\%}$ & $\mathbf{76.83\%}$\\
         $30\%$ & $76.18\%$ & $\mathbf{76.23\%}$ & $\mathbf{76.55\%}$\\
         \hline
    \end{tabular}
    }
    \end{minipage}
    \caption{Comparison between pruning during training with EPI and pruning a pre-trained model. 
    }
    \vspace{-4mm}
    \label{tab:pruning-compare}
\end{table}

\section{Conclusions}
We have introduced an approach to automatically determine when pruning can be performed during training without affecting the final accuracy and with the additional constraint of doing so as early as possible. To this end, we have proposed a policy based on Early Pruning Indicator (EPI), a metric to measure the stability of the sub-network structure. Our experiments on multiple pruning algorithms and pruning ratios have demonstrated the benefits of our method to reduce the accuracy drop when pruning a network and observe a significant reduction in training time.



{\small
\bibliographystyle{ieee_fullname}
\bibliography{egbib}
}

\clearpage
\newpage
\begin{appendices}

We provide more experimental details in the following sections.

\section{Additional EPI-guided pruning results}
\label{epi-guided-pruning-results}
We perform EPI-guided pruning, with gradient-based criterion, on ResNet50, ResNet34 and MobileNetV1 with different prune ratios and get the final accuracy as shown in Tab.~\ref{tab:gradient-based-top1-acc-policy-compare}. It can be observed that over all pruning ratios, EPI demonstrates superior capability over prior methods in achieving similar performance to oracle while pushing the start of pruning earlier into training.

Similarly, for magnitude-based pruning in Tab.~\ref{tab:magnitude-based-top1-acc-policy-compare} we show the results of pruning with the magnitude universal threshold. In this case, as before, EPI-guided pruning performs better than heuristic pruning. Note that in magnitude-based pruning, pruning at initialization (heuristically pruning at epoch $0$) tends to lead to non-trainable networks given a more challenging task amid model compactness. 

\begin{table}[t]
    \begin{minipage}{\columnwidth}
    \centering
    \resizebox{.95\columnwidth}{!}{
    \begin{tabular}{|c|c|c|c|c|c|}
        \cline{2-6}
        \multicolumn{1}{c|}{} & \tabincell{c}{Pruning\\ratio} & \tabincell{c}{Oracle\\estimate} & \tabincell{c}{Init.\\\cite{de2020force,lee2018snip}} & \tabincell{c}{Pre-def\\\cite{he2020learning}} & EPI (ours)\\
        \hline
        \multirow{9}{*}{\begin{sideways}ResNet50\end{sideways}} & $10\%$ & $76.90$ & $76.80$ & $76.74$ & $\mathbf{76.83} _{(5)}$\\
        & $20\%$ & $76.37$ & $76.23$ & $76.32$ & $\mathbf{76.34} _{(9)}$\\
        & $30\%$ & $75.78$ & $74.88$ & $75.60$ & $\mathbf{75.75} _{(11)}$\\
        & $40\%$ & $74.93$ & $73.73$ & $74.85$ & $\mathbf{74.91} _{(12)}$\\
        & $50\%$ & $73.89$ & $71.92$ & $73.62$ & $\mathbf{73.84} _{(12)}$\\
        & $60\%$ & $71.97$ & $70.04$ & $\mathbf{71.92}$ & $71.79 _{(12)}$\\
        & $70\%$ & $69.00$ & $66.65$ & $68.49$ & $\mathbf{68.92} _{(13)}$\\
        & $80\%$ & $63.77$ & $60.65$ & $62.68$ & $\mathbf{63.40} _{(17)}$\\
        & $90\%$ & $61.65$ & $48.73$ & $\mathbf{61.65}$ & $\mathbf{61.65} _{(30)}$\\
        \hline
        \hline
        \multirow{5}{*}{\begin{sideways}ResNet34\end{sideways}} & $10\%$ & $73.76$ & $73.58$ & $\mathbf{73.67}$ & $73.59 _{(5)}$\\
        & $20\%$ & $72.56$ & $72.24$ & $\mathbf{72.42}$ & $72.31 _{(10)}$\\
        & $30\%$ & $70.63$ & $\textbf{70.53}$ & $70.24$ & $70.28 _{(13)}$\\
        & $40\%$ & $68.20$ & $\mathbf{68.20}$ & $68.04$ & $68.00 _{(7)}$\\
        & $50\%$ & $65.54$ & $65.52$ & $65.22$ & $\mathbf{65.53} _{(13)}$\\
        \hline
        \hline
        \multirow{5}{*}{\begin{sideways}MobileNetV1\end{sideways}} & $10\%$ & $72.50$ & $72.34$ & $\mathbf{72.40}$ & $72.27 _{(5)}$\\
        & $20\%$ & $71.66$ & $71.48$ & $\mathbf{71.59}$ & $\mathbf{71.59 _{(5)}}$\\
        & $30\%$ & $70.69$ & $70.38$ & $\mathbf{70.56}$ & $70.49 _{(6)}$\\
        & $40\%$ & $69.22$ & $\mathbf{69.09}$ & $69.07$ & $69.05 _{(12)}$\\
        & $50\%$ & $67.25$ & $67.10$ & $\mathbf{67.15}$ & $67.03 _{(5)}$\\
        \hline
        \hline
        \multicolumn{2}{|c|}{\tabincell{c}{all nets avg acc drop}} & -- & $1.378$ & $0.214$ & $\mathbf{0.142}$ \\
        \hline
        \multicolumn{2}{|c|}{\tabincell{c}{automatic}} & \xmark & $ \checkmark $ & \xmark & $ \checkmark $ \\
        \hline
    \end{tabular}
    }
    \end{minipage}
    \caption{The detailed top1 accuracy (in \%) of the network with \textbf{gradient}-based neuron pruning, under different policies as in Section 4.2 of the main paper. Each reported value under our proposed EPI policy is in the format of $\text{[top1 acc]}_{(\text{prune epoch})}$. The summarize of the accuracy drop is the averaged accuracy drop (in \%) compared to the oracle method.}
    \label{tab:gradient-based-top1-acc-policy-compare}
\end{table}

\begin{table}[t]
    \begin{minipage}{\columnwidth}
    \centering
    \resizebox{.95\columnwidth}{!}
    {
    \begin{tabular}{|c|c|c|c|c|c|}
        \cline{2-6}
        \multicolumn{1}{c|}{} & \tabincell{c}{prune\\ratio} & \tabincell{c}{Oracle\\estimate} & \tabincell{c}{Init.\\\cite{de2020force,lee2018snip}} & \tabincell{c}{Pre-def\\\cite{he2020learning}} & EPI (ours)\\
        \hline
        \multirow{5}{*}{\begin{sideways}ResNet50\end{sideways}} & $10\%$ & $76.9$ & $74.50$ & $\mathbf{76.90}$ & $76.76 _{(15)}$ \\
        & $20\%$ & $76.25$ & $72.71$ & $\mathbf{76.13}$ & $76.12 _{(17)}$\\
        & $30\%$ & $75.56$ & $70.69$ & $75.42$ & $\mathbf{75.46} _{(18)}$\\
        & $40\%$ & $74.53$ & -- & $74.46$ & $\mathbf{74.49} _{(20)}$\\
        & $50\%$ & $72.95$ & -- & $72.71$ & $\mathbf{72.90} _{(26)}$\\
        \hline
        \hline
        \multirow{5}{*}{\begin{sideways}ResNet34\end{sideways}} & $10\%$ & $73.78$ & $73.49$ & $\mathbf{73.71}$ & $73.64 _{(15)}$\\
        & $20\%$ & $72.97$ & $72.20$ & $72.78$ & $\mathbf{72.86} _{(16)}$\\
        & $30\%$ & $71.63$ & $71.40$ & $71.53$ & $\mathbf{71.58} _{(19)}$\\
        & $40\%$ & $69.94$ & $67.90$ & $69.78$ & $\mathbf{69.84 _{(21)}}$\\
        & $50\%$ & $67.42$ & -- & $66.19$ & $\mathbf{66.97 _{(24)}}$\\
        \hline
        \hline
        \multirow{5}{*}{\begin{sideways}MobileNetV1\end{sideways}} & $10\%$ & $71.87$ & -- & $71.65$ & $\mathbf{71.87} _{(5)}$\\
        & $20\%$ & $70.59$ & -- & $70.34$ & $\mathbf{70.58} _{(9)}$\\
        & $30\%$ & $68.91$ & -- & $\mathbf{68.88}$ & $68.52 _{(6)}$\\
        & $40\%$ & $66.60$ & -- & $\mathbf{66.60}$ & $66.50 _{(15)}$\\
        & $50\%$ & $63.28$ & -- & $\mathbf{63.11}$ & $\mathbf{63.11} _{(30)}$\\
        \hline
        \hline
        \multicolumn{2}{|c|}{\tabincell{c}{all nets avg acc drop}} & -- & $2.023$\footnote{\label{note1} is averaged over the trainable pruning results.} & $0.204$ & $\mathbf{0.132}$ \\
        \hline
        \multicolumn{2}{|c|}{\tabincell{c}{automatic}} & \xmark & $\checkmark$ & \xmark & $\checkmark$ \\
        \hline
    \end{tabular}
    }
    \end{minipage}
    \caption{The detailed top1 accuracy (in \%) of the network with \textbf{magnitude}-based neuron pruning, under different policies as in Section 4.2 of the main paper. ``--'' refers to pruning leads to not-trainable network. Each reported value under our proposed EPI policy is in the format of $\text{[top1 acc]}_{(\text{prune epoch})}$. The summarize of the accuracy drop is the averaged accuracy drop (in \%) compared to the oracle method.}
    \label{tab:magnitude-based-top1-acc-policy-compare}
\end{table}

\section{Stability analysis for MobileNetV1}
\begin{figure}[h]
    \centering
    \begin{tabular}{cc}
        \includegraphics[width=0.45\columnwidth]{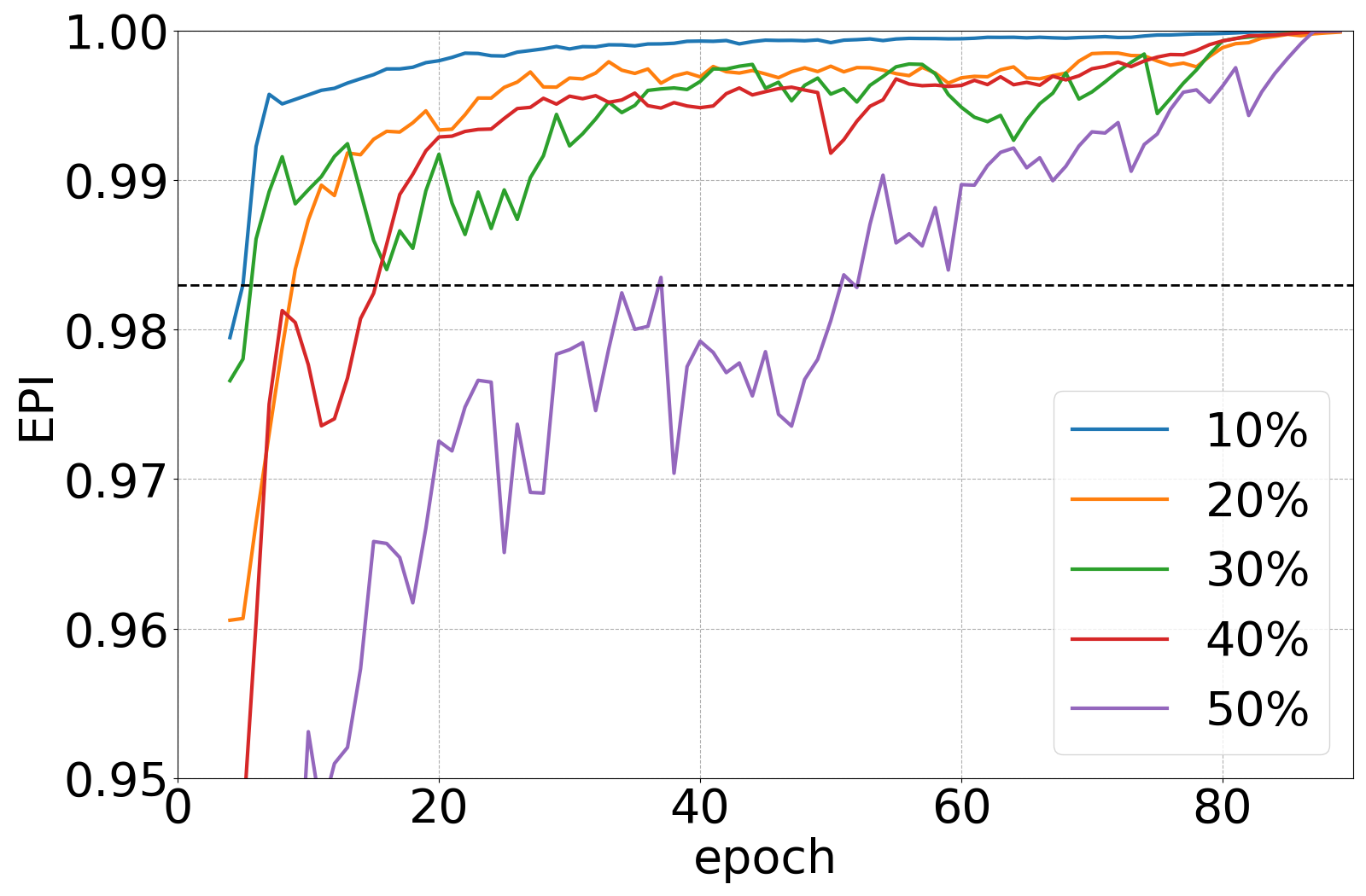} & \includegraphics[width=0.45\columnwidth]{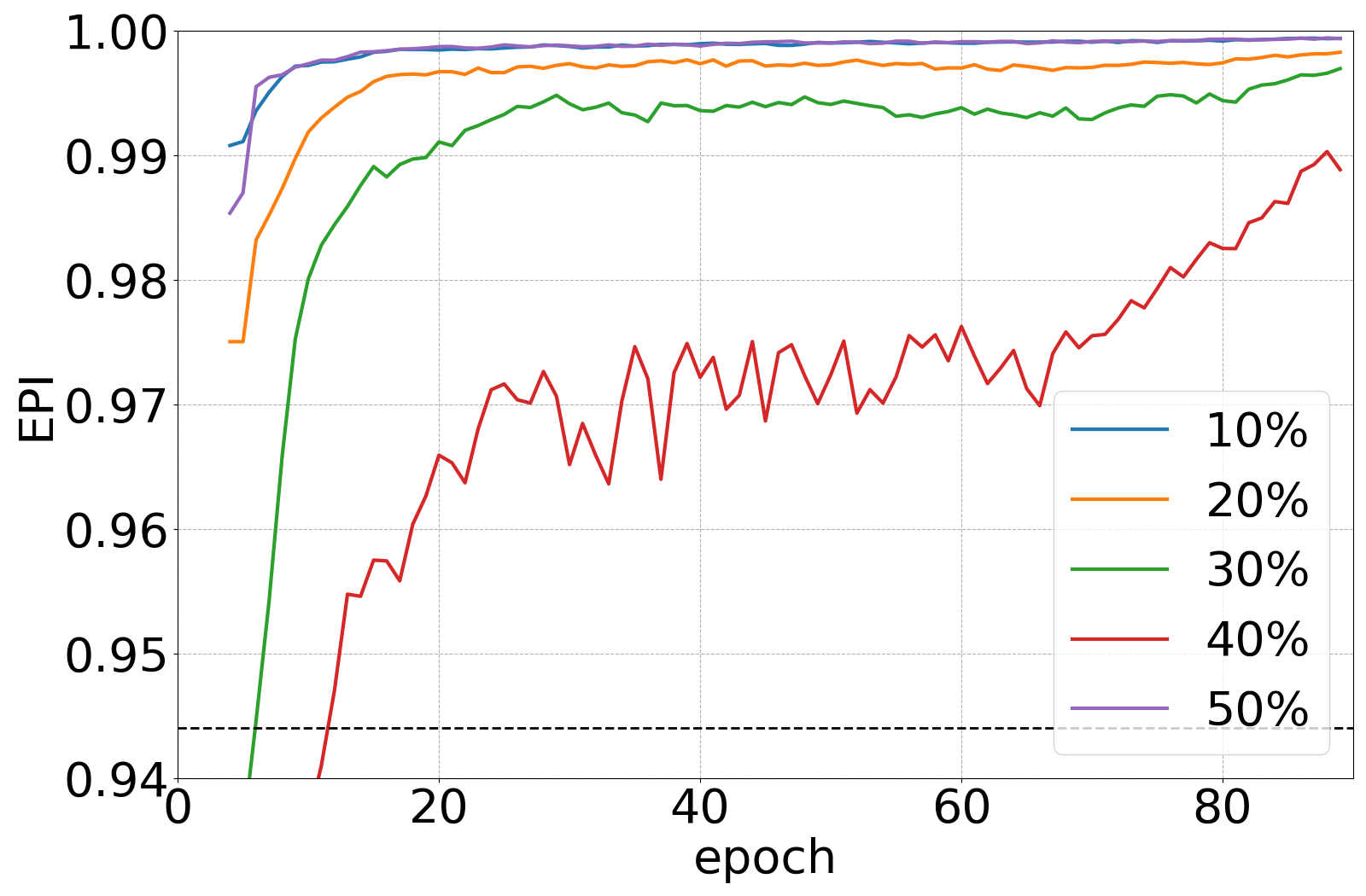} \\
        (a) magnitude-based EPI & (b) gradient-based EPI
    \end{tabular}
    \caption{Structure stability analysis for MobileNetV1 for magnitude-based (a) and gradient-based (b) pruning with different ratios. Dashed line in black shows the EPI threshold.}
    \label{fig:mobilenet-epi}
\end{figure}

We provide here additional stability analysis results for a MobileNetV1 architecture. As in Section 4.5 of the main paper, the goal is to demonstrate that the sub-network architecture varies significantly during the early stage of training and then slowly converges to the final architecture as the training progresses. In this case, we train a MobileNetV1 to convergence and, in the process, compute the EPI value for different prune ratios. Fig.~\ref{fig:mobilenet-epi}(a) and (b) show the EPI curves with magnitude-based and gradient-based, respectively. Consistent with our main paper results, the stability indicator increases rapidly in the early stage of training and continues increasing steadily for later training stages. 

\section{Comparison to other similarity criteria}
\begin{figure}[!t]
    \centering
    \includegraphics[width=0.6\columnwidth]{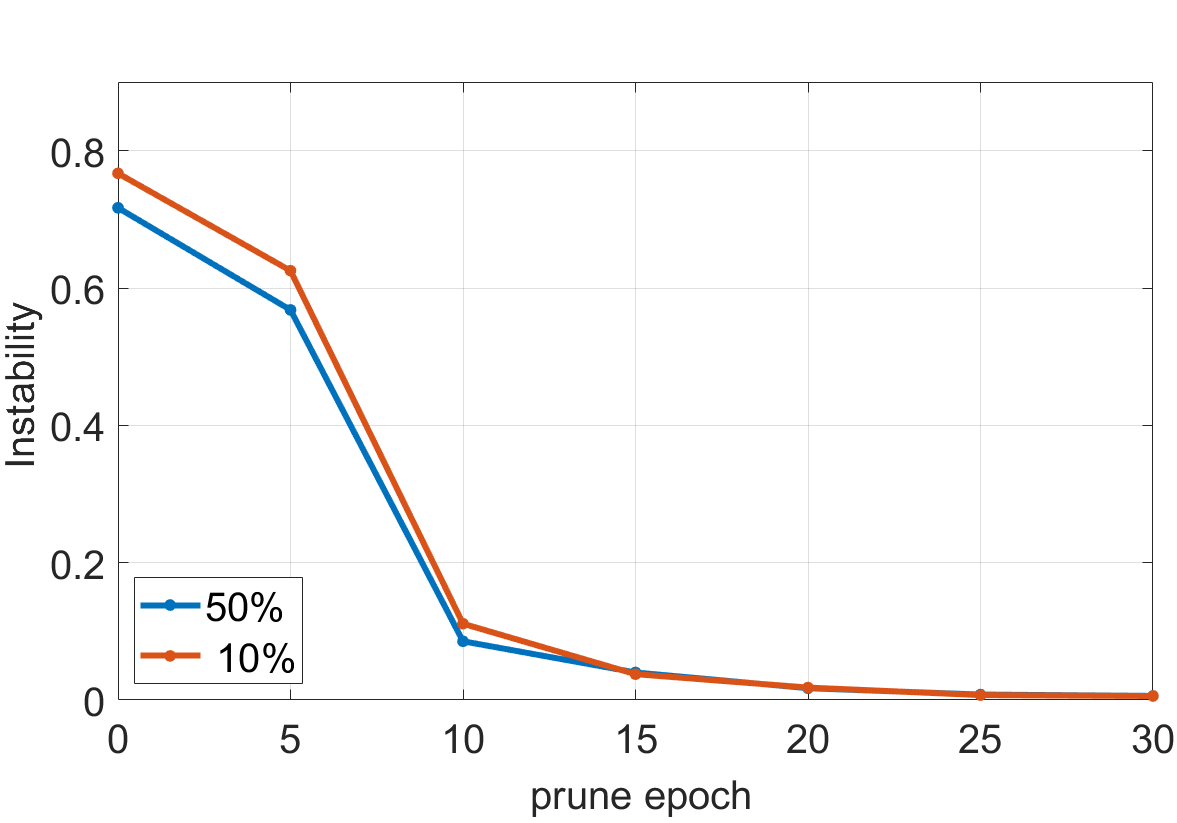}
    \caption{Instability \cite{frankle2019linear} of the networks obtained by pruning at different epochs. In this case, the prune ratio is involved, but the curves are not distinguishable among different prune ratios. Moreover, such instability can only be calculated after training.}
    \label{fig:rankComparisons}
\end{figure}

\begin{figure}[!t]
    \centering
    \includegraphics[width=0.6\columnwidth]{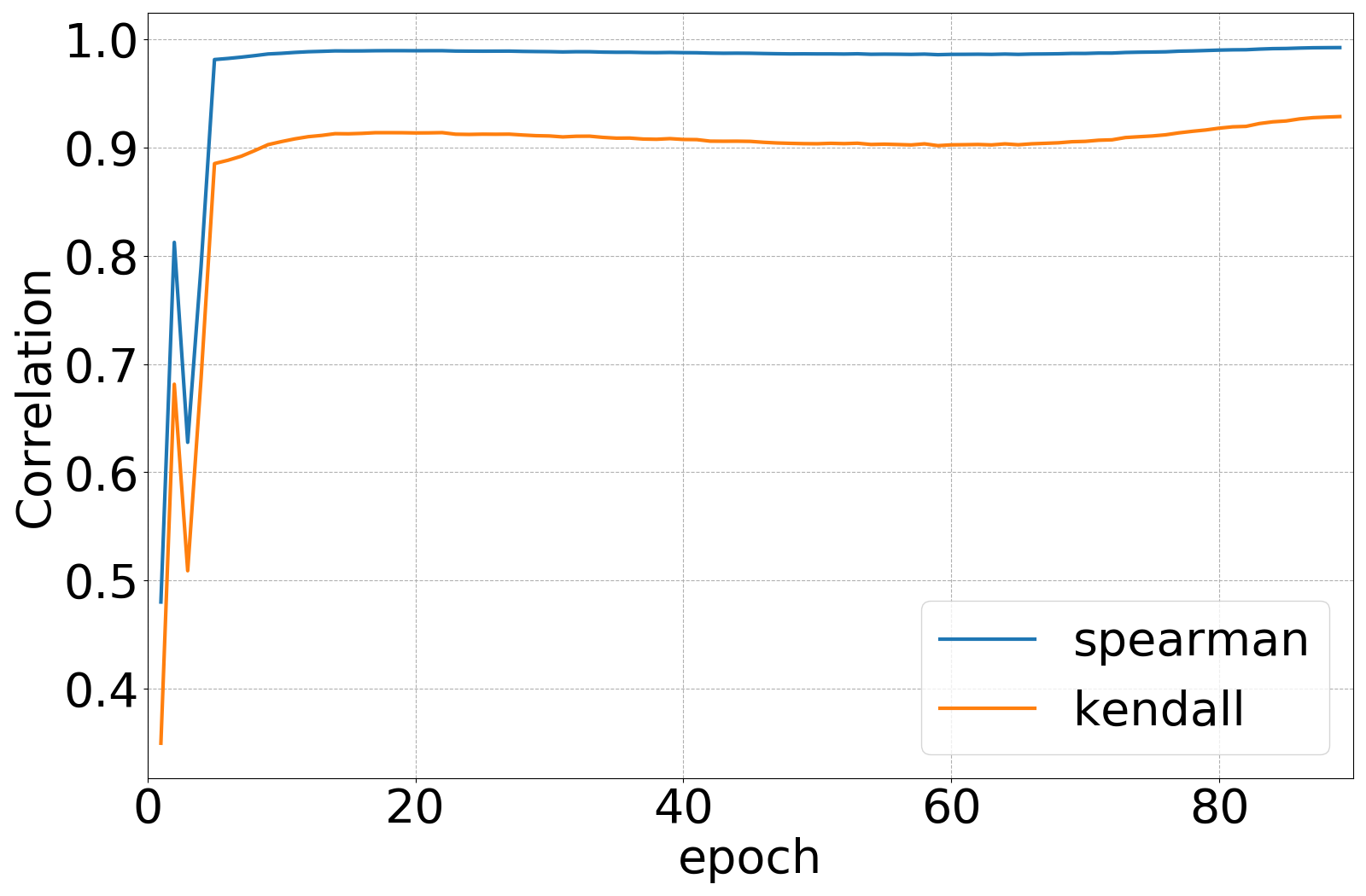}
    \caption{Kendall and Spearman correlation calculated based on gradient-based neuron metric ranking. Each value is the calculated between epoch $x$ and epoch $x-1$. With rank correlation, no prune ratio would be involved. That means if we set a threshold to the correlation value, then for whatever prune ratio, we need to perform pruning at the same epoch.}
    \label{fig:rankComparisons2}
\end{figure}

In this experiment, we aim at comparing our proposed EPI to other ranking criteria to measure the difference between two network structures. To this end, we consider two ranking correlation measures (spearman and Kendall tau) and the instability measure proposed in~\cite{frankle2019linear}. Ranking correlation approaches directly measure the difference in the neuron ranking. These ranking correlation measures require all the neurons in the architecture and can not discriminate between different pruning ratios. Instability focuses on identical architectures trained with different SGD noise and is computed after training is completed. Fig.~\ref{fig:rankComparisons} and Fig.~\ref{fig:rankComparisons2} show the results for this experiment on ResNet50. As expected, using other measures, we not only can not distinguish different pruning ratios but also do not have enough discriminative power even during the early stages of training. In contrast, as our approach focuses only on the architecture changes can, as shown in Fig. 6 in the main paper, provide better insights into the stability of the network. 


\section{EPI-guided Pruning on Object Detection}
\begin{figure}
    \centering
    \begin{tabular}{cc}
        \includegraphics[width=0.45\columnwidth]{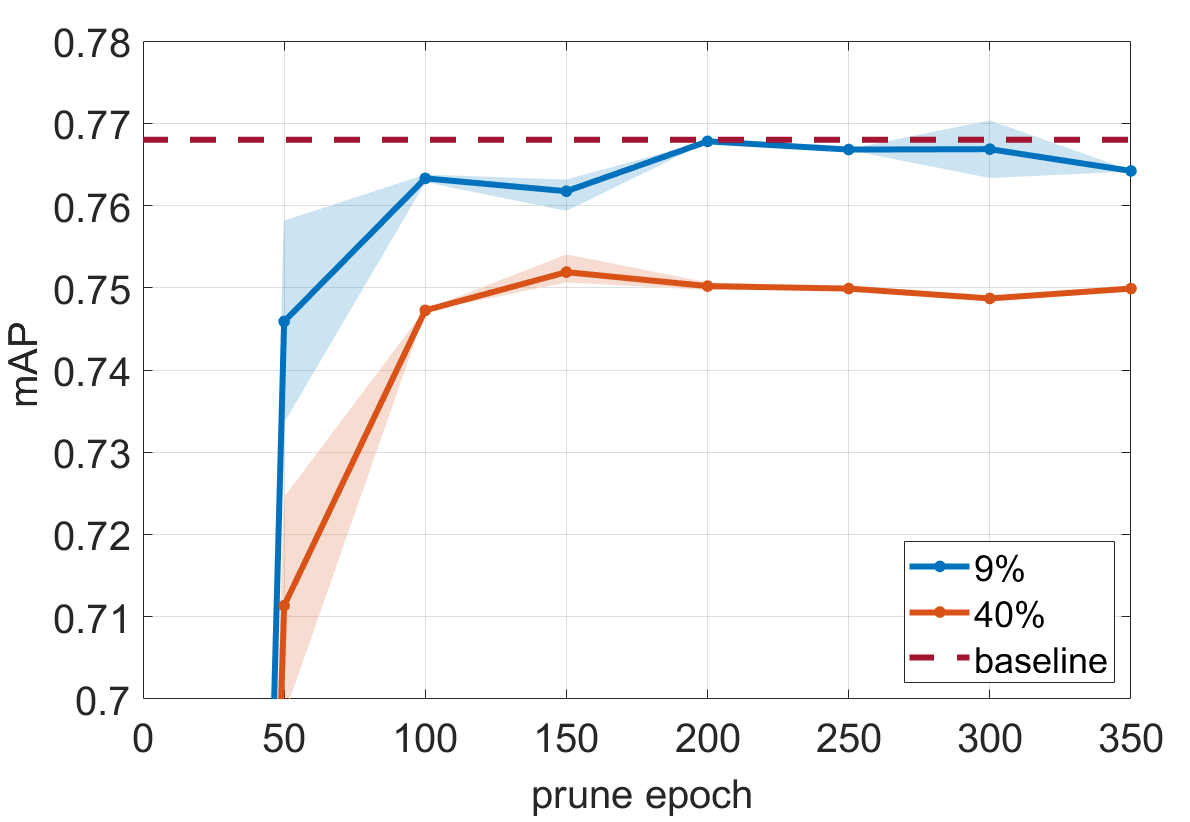}&
        \includegraphics[width=0.45\columnwidth]{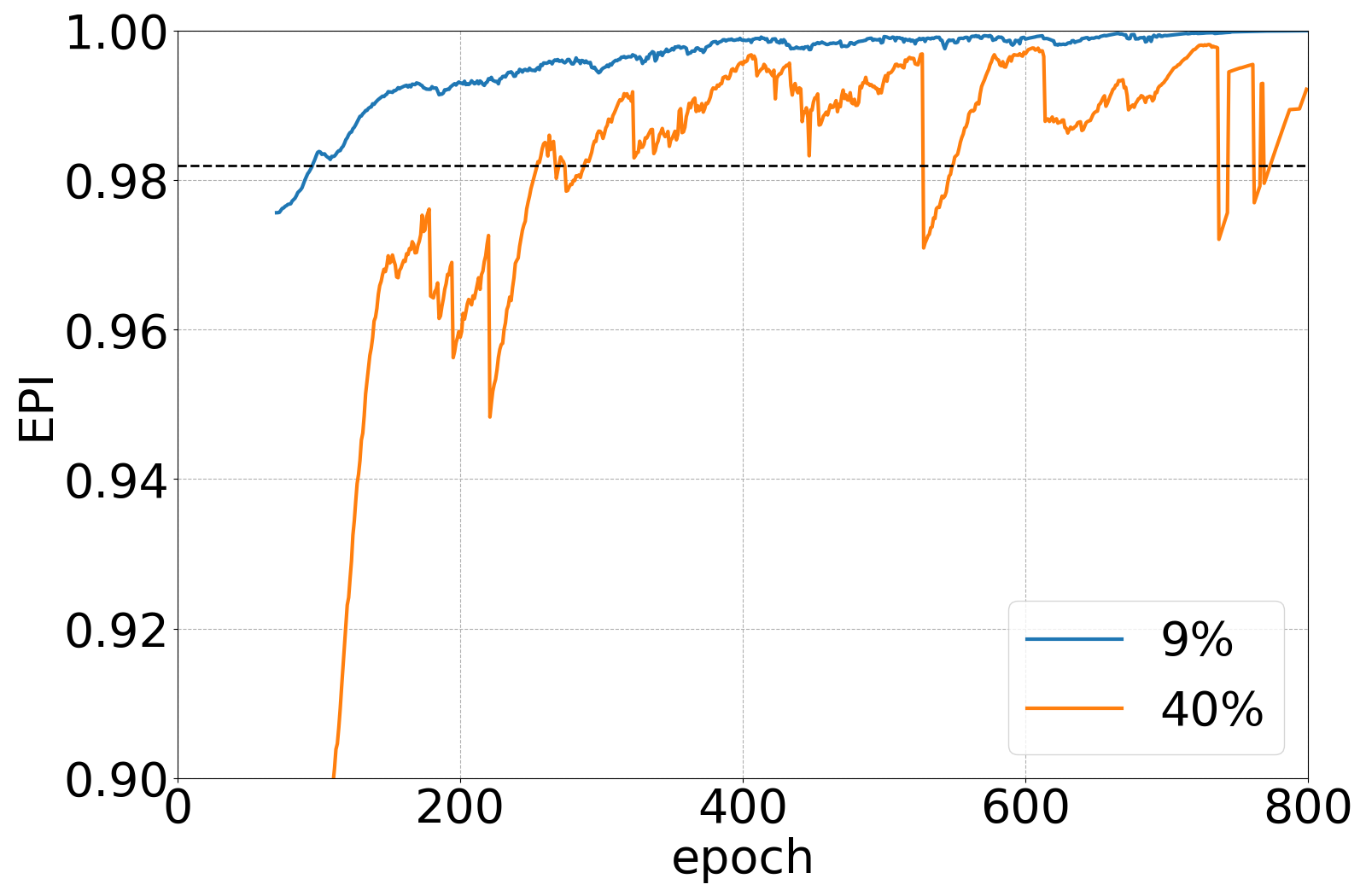}\\
        (a)&(b)
    \end{tabular}
    \caption{Object detection using SSD300-RN34 on Pascal VOC. (a) The magnitude-based pruning result and (b) Structure stability analysis (EPI curve) for SSD300-RN34. Dashed line in black shows the EPI threshold.}
    \label{fig:ssd300-prune}
\end{figure}
In this experiment, we extend our proposed EPI policy to object detection. We use a Single Shot multibox Detector (SSD)~\cite{liu2016ssd} as our detection network, with a ResNet34 backbone and an input size resolution of $300\times300$. We run the experiments on PASCAL VOC07+12 (union of VOC2007 and VOC2012)~\cite{everingham2010pascal}. 

For training, we use PyTorch Distributed Data Parallel and mixed precision. We train the model for $800$ epochs in total, with an individual batch per GPU of $128$. The learning rate is warmed up linearly to $8e-3$ in the first $50$ epochs, remains at the maximum value until epoch $600$, and decays every $50$ epochs. As an upper bound and baseline, we consider the accuracy of the unpruned model where we obtained $76.8\%$ mAP. 

We test magnitude-based pruning with $9\%$ and $40\%$ pruning ratio and compare the performance to grid-search pruning with $50$ epochs. Fig.~\ref{fig:ssd300-prune}(a) shows the results for this experiment. As we can see, pruning too early leads to large accuracy drops. However, if we delay the pruning epoch, especially with a lower pruning ratio, we do achieve on-par accuracies with the upper bound. 

We also compute the EPI value, see Eq.(6) in the main paper, for these two pruning ratios. In this case, different from our classification set up, we calculate the sub-network structure similarity among the past $50$ epochs, \ie, $r=50$. Fig.~\ref{fig:ssd300-prune}(b) shows the EPI curve for this experiment. As shown, the EPI value increases rapidly at the early stage of training and then increases gradually as the training progresses. The tendency is consistent with or pruning results in Fig.~\ref{fig:ssd300-prune}(a). As in our previous experiments, we set an EPI threshold to the magnitude universal threshold value $\tau=0.983$. Given this threshold, for this architecture, using EPI-guided pruning leads to pruning in the $96$th epoch for $9\%$ pruning ratio and pruning in the $255$th for a pruning ratio of $40\%$. The mAp drop with respect to the grid-search result is $0.589\%$ and $0.212\%$ mAP respectively.

\end{appendices}

\end{document}